\documentclass[preprint,3p,times,twocolumn]{elsarticle} % default

\usepackage{amssymb}
\usepackage{lineno}
\usepackage{amsmath}

\journal{Renewable Energy}

\begin{document}

\begin{frontmatter}

%% Title, authors and addresses

%% use the tnoteref command within \title for footnotes;
%% use the tnotetext command for theassociated footnote;
%% use the fnref command within \author or \address for footnotes;
%% use the fntext command for theassociated footnote;
%% use the corref command within \author for corresponding author footnotes;
%% use the cortext command for theassociated footnote;
%% use the ead command for the email address,
%% and the form \ead[url] for the home page:
%% \title{Title\tnoteref{label1}}
%% \tnotetext[label1]{}
%% \author{Name\corref{cor1}\fnref{label2}}
%% \ead{email address}
%% \ead[url]{home page}
%% \fntext[label2]{}
%% \cortext[cor1]{}
%% \affiliation{organization={},
%%             addressline={},
%%             city={},
%%             postcode={},
%%             state={},
%%             country={}}
%% \fntext[label3]{}

\title{Omnivision forecasting: combining satellite observations with sky images for improved intra-hour solar energy predictions}

%\title{Omnivision forecasting: predicting intra-hour solar energy from satellite and sky images}
%\title{Vision-based short-term solar energy forecasting from sky and satellite images}
%\title{Multi-view Solar Energy Forecasting}
%\title{Long Short Term Multi-view Solar Energy Forecasting}
%\title{Sky and satellite video processing for short-term solar energy forecasting}
%\title{Short-term irradiance forecasting from sky and satellite images with machine vision}

%\author{Quentin Paletta}
%\author[label2]{Guillaume Arbod}
%\author{Joan Lasenby}

\author[label1,label2]{Quentin Paletta\corref{cor1}}
\author[label2]{Guillaume Arbod}
\author[label1]{Joan Lasenby}

\address[label1]{Department of Engineering, University of Cambridge, UK}
\address[label2]{ENGIE Lab CRIGEN, France}

%\affiliation[label1]{organization={University of Cambridge},
%            addressline={Trumpinton street}, 
%            city={Cambridge},
%            postcode={CB3}, 
%            state={Cambridgeshire},
%            country={UK}}
            
%\affiliation[label2]{organization={Lab CRIGEN, Engie},%Department and Organization
%            addressline={}, 
%            city={},
%            postcode={}, 
%            state={},
%            country={France}}

\cortext[cor1]{E-mail address: qp208@cam.ac.uk (Quentin Paletta)}

\begin{abstract}
%% Text of abstract

Integration of intermittent renewable energy sources into electric grids in large proportions is challenging. A well-established approach aimed at addressing this difficulty involves the anticipation of the upcoming energy supply variability to adapt the response of the grid. In solar energy, short-term changes in electricity production caused by occluding clouds can be predicted at different time scales from all-sky cameras (up to 30-min ahead) and satellite observations (up to 6h ahead). In this study, we integrate these two complementary points of view on the cloud cover in a single machine learning framework to improve intra-hour (up to 60-min ahead) irradiance forecasting. Both deterministic and probabilistic predictions are evaluated in different weather conditions (clear-sky, cloudy, overcast) and with different input configurations (sky images, satellite observations and/or past irradiance values). Our results show that the hybrid model benefits predictions in clear-sky conditions and improves longer-term forecasting. This study lays the groundwork for future novel approaches of combining sky images and satellite observations in a single learning framework to advance solar nowcasting.

%The proposed deep learning approach is shown to outperform statistical models and traditional computer vision techniques.

\end{abstract}

%%Graphical abstract
%\begin{graphicalabstract}
%\includegraphics{grabs}
%\end{graphicalabstract}

%%Research highlights
%\begin{highlights}
%\item We combine both sky and satellite imagery to improve short-term irradiance predictions
%\item We adapt current state-of-the-art deep learning architecture to this multi-view vision task
%\end{highlights}

\begin{keyword}
%% keywords here, in the form: keyword \sep keyword

%% PACS codes here, in the form: \PACS code \sep code

%% MSC codes here, in the form: \MSC code \sep code
%% or \MSC[2008] code \sep code (2000 is the default)

Solar energy \sep Nowcasting \sep Computer vision \sep Deep learning \sep Satellite observations \sep Sky images

\end{keyword}

\end{frontmatter}

%% \linenumbers

%% main text
\section{Introduction}
\label{introduction}

Solar energy is expected to become a key contributor to the current energy transition towards spread and low carbon electricity. Despite a rapid increase of its globally installed capacity (\citet{InternationalEnergyAgencyIEA2018}), a range of technical obstacles such as land requirement or weather-dependent intermittency are still limiting its broad utilisation (\citet{capellan-perezAssessingVulnerabilitiesLimits2017}). In addition to improving storage technologies (\citet{koenLowtemperatureGlideCycle2021}), accurate energy supply forecasting at different spatio-temporal scales would facilitate power grid balancing (\citet{inmanSolarForecastingMethods2013, tawnReviewVeryShortterm2022a, impramChallengesRenewableEnergy2020}). To that end, local and remote sensing offer an opportunity to measure weather variability and model its impact on solar energy generation (\citet{yangHistoryTrendsSolar2018}). 

%\vspace{0.5\baselineskip}
According to~\citet{IEA_renewable_energy_market_update}, utility-scale solar power accounts for an increasing share of the yearly capacity additions. Contrary to residential photovoltaic electricity production whose spread out spatial distribution lessens the effects of local intermittency caused by clouds, output of large solar farms can be heavily impacted by local cloud cover changes. In addition to numerical weather models and statistical techniques, vision-based forecasting has strongly contributed to improve local solar energy predictions (\citet{antonanzasReviewPhotovoltaicPower2016a}). In particular, computer vision approaches based on observations of the cloud cover can better anticipate critical events causing large production shifts.

\subsection{Hemispherical cameras}
\label{hemispherical_cameras}
%\vspace{0.5\baselineskip}
At a short-term scale, \emph{in situ} observations of the cloud cover from hemispherical sky-cameras offer a high spatiotemporal resolution (\citet{peng3DCloudDetection2015a, blancShorttermForecastingHigh2017a}): the field of view of a single camera reaches about 1.5 $\text{km}^2$ and its frequency of acquiring data ranges from a few seconds to several minutes. Well suited for solar farm monitoring, hemispherical cameras can be used to estimate the local irradiance map based on observable properties of the cloud cover (\citet{nouriDeterminationCloudTransmittance2019}) and its estimated spatial configuration (\citet{kuhnDeterminationOptimalCamera2019, nouriNowcastingDNIMaps2018}). Methods based on 2D cross-correlation or optical flow enable cloud tracking (\citet{Wood-Bradley2012, quesada-ruizCloudtrackingMethodologyIntrahour2014a}) and cloud motion prediction (\citet{nouriHybridSolarIrradiance}) from a sequence of sky images.

A distinct computer vision approach involves the application of neural networks to the forecasting task. In particular, convolutional neural networks (CNNs) have been trained to correlate an image with the corresponding solar flux (\citet{sunSolarPVOutput2018, insafGlobalHorizontalIrradiance2021}) or to forecast irradiance from a sequence of sky images (\citet{zhangDeepPhotovoltaicNowcasting2018}). Trained convolutional architectures can recognise specific cloud patterns and adjust their prediction accordingly~\cite{wenDeepLearningBased2021, palettaConvolutionalNeuralNetworks2020}. Despite promising results, current architectures still have difficulties in breaking the \textit{persistence barrier} to anticipate critical events such as a cloud obstructing the sun (\citet{palettaBenchmarkingDeepLearning2021c}). These events are both more challenging to predict (cloud cover changes are ruled by complex physical phenomena) and rare (the shorter the forecast horizon, the more unbalanced is a dataset towards samples with little irradiance change). To improve the predictability of critical events and decrease the time lag associated with predictions, novel temporal encoders have been developed and models are being supervised during training by a video prediction task (\citet{leguenDeepPhysicalModel2020a, palettaECLIPSEEnvisioningCloud2021}). Although sampling strategies have been implemented to address the unbalance of the dataset, limited gains on the forecasting task were observed (\citet{nieResamplingDataAugmentation2021}).

\subsection{Satellite imagery}
\label{satellite_imagery}

Solar energy forecasting from satellite images is well suited for longer-term forecasts (30-min to 6h) because of its lower spatial (about 1$\text{km}^2$) and temporal (5 to 15-min) resolutions as well as its larger field of view covering most of the planet (\citet{inmanSolarForecastingMethods2013}). Similarly to sky image analysis, optical and physical properties of clouds can be exploited to localise clouds and estimate their transmittance to solar flux (\citet{cebecauerHIGHPERFORMANCEMSG2010, waldUSERGUIDEMACCRAD2015}). The cloud index, also called effective cloud albedo, used to estimate the surface solar irradiance (\citet{muellerRoleEffectiveCloud2011}), can be derived by subtracting the ground albedo from the current cloudy image (see Section~\ref{satellite_images} and Figure~\ref{fig:satellite} for more details). Information about the spatio-temporal dynamics of the cloud cover is obtained by comparing subsequent images to derive the so-called cloud motion vectors (CMVs) using block matching (\citet{dazhiBlockMatchingAlgorithms2013}) or optical flow (\citet{urbichSeamlessSolarRadiation2019}). These techniques primarily rely on the assumption that cloud cover changes are only explained by the local translation of clouds. Thus, the formation or dissipation of clouds are not taken into account.

Common solar energy forecasting methods derived from surface irradiance maps range from auto-regressive models (\citet{dambrevilleVeryShortTerm2014}), machine learning algorithms (\citet{jangSolarPowerPrediction2016a, larsonDirectPowerOutput2018}) and artificial neural networks (\citet{voyantTimeSeriesModeling2014, srivastavaComparativeStudyLSTM2018, lagoShorttermForecastingSolar2018}). The reader may refer to (\citet{blancShorttermSolarPower2017}) for more details.

More recently, convolutional neural networks have been used to predict future frames or future irradiance levels from a sequence of past satellite images (\citet{perezDeepLearningModel2021b, siPhotovoltaicPowerForecast2021a, nielsenIrradianceNetSpatiotemporalDeep2021a}). Contrary to sky images which correlate well with their corresponding irradiance level, raw satellite observations are often supplemented by additional information on the current solar irradiance value to improve the performance. \citet{perezDeepLearningModel2021b} do not rely on ground measurements but use physics-based surface solar irradiance maps as a 2D input to a CNN model. However, they also evaluate the potential benefit of integrating past irradiance measurements and corresponding clear-sky irradiance levels to calibrate the model. A follow-up study by \citet{nielsenIrradianceNetSpatiotemporalDeep2021a} uses recurrent layers (ConvLSTM) to improve temporal feature extraction from a sequence of past satellite observations. They chose to focus on the cloud dynamics by considering cloud index maps which can be easily derived without external parameters (\citet{muellerRoleEffectiveCloud2011}). Similarly to~\citet{sonderbyMetNetNeuralWeather2020}, their model `\textit{IrradianceNet}' is augmented to provide probabilistic forecasts by discretizing the range of effective cloud albedo. To address the parallax phenomenon caused by the non-alignment of the satellite with the direct incoming irradiance, input satellite images can be centred on the area of the cloud cover which is between the power plant and the sun instead of centring it on the power plant (\citet{siPhotovoltaicPowerForecast2021a}).

\subsection{Irradiance forecasting from sky and satellite images}
\label{satellite_andsky_imagery}

So far, short-term vision-based solar forecasting from sky images or satellite observations have progressed independently, mainly because of the different spatio-temporal resolutions of these two data sources. For this reason, approaches involving sky images focus on very short-term irradiance forecasting (up to 30-min) while those using satellite images address longer-term predictions (from 30-min to several hours). However, due to the similarity of the computer vision tasks involved (cloud segmentation, cloud tracking, solar flux prediction, etc.), some of the methods discussed above have been experimented with in both contexts (optical flow, block matching, image segmentation using neural networks, etc.).

Some articles have tried to compare both forecasting approaches (from sky or satellite images) independently in different sky conditions and different forecast horizons to evaluate the conditions in which one prevails over the other (\citet{alonsoShortMediumtermCloudiness2014, rodriguez-benitezAssessmentNewSolar2021}). However, to the best of our knowledge, the only work proposing an actual data fusion approach was done by~\citet{Vallance2018} in his doctorate dissertation. The main idea of his work is to localise clouds in three dimensions using voxel carving (\citet{Oberlander2015}) from both satellite and sky-camera observations as two complementary points of view on the cloud cover. However, his method was tested on a single test day and has not yet been evaluate on a bigger dataset.

\vspace{0.5\baselineskip}

\textbf{Contributions} In this study, a hybrid deep learning (DL) model for intra-hour irradiance nowcasting is proposed. The model is trained to predict future cloud index maps and local irradiance measurements from both ground-taken sky images (\citet{sirta}) and satellite images (\citet{eumetsatorganizationEUMETSATEuropeanOrganisation1991}). Both deterministic and probabilistic approaches are compared and evaluated quantitatively and qualitatively using standard metrics showing the benefit of integrating both sources of data from a 25-min lead time. In particular, the specific benefit of the different sources of data (irradiance measurements, sky images, satellite observations) are evaluated in different weather conditions (clear-sky, broken-sky and overcast). The gain of the video prediction task on the quality of the solar forecast is also quantified.

\vspace{0.5\baselineskip}

The dataset used in this study is presented in Section~\ref{data}. Section~\ref{methodology} details different aspects of the methodology followed in this study. Results and limitations are outlined and discussed in the following Sections~\ref{results} and~\ref{discussion}.

\section{Data}
\label{data}

\subsection{Irradiance}

The solar irradiance measurements used in this study originate from Sirta's lab in France (\citet{sirta}). They were collected over three years from 2017 to 2019. The global horizontal irradiance (GHI) was measured on site with a pyranometer and reported as its per minute average. Samples corresponding to a solar zenith angle greater than 80° were discarded from the experiment.

\subsection{Sky images}

\begin{figure}%[H]%[ht!]%[h!] 
\centering
\begin{minipage}[b]{0.23\textwidth}
    \includegraphics[width=0.995\textwidth]{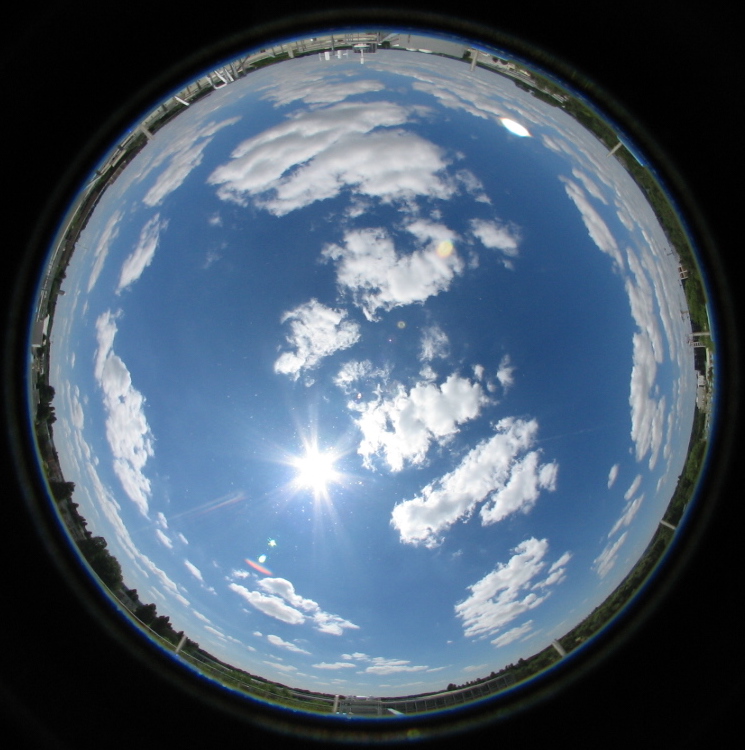}
  \end{minipage}
  \begin{minipage}[b]{0.23\textwidth}
    \includegraphics[decodearray={0.1 1.8}, width=1\textwidth]{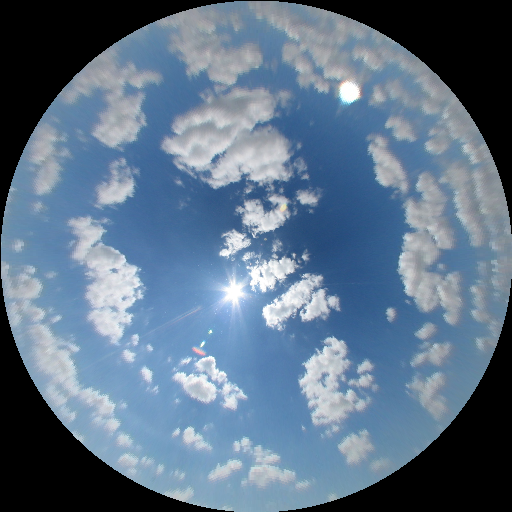}
  \end{minipage}
%\vspace{-1.2\baselineskip}
\caption{Raw and undistorted sky image taken by an hemispherical sky-camera. Notice how the relative size of distant clouds increases in the projected image.}
\label{fig:sky_images}
\end{figure}

Sky images were taken by an EKO SRF-02 all-sky camera located close to the pyranometer used to measure irradiance (\citet{sirta}). The temporal resolution ranges from 1-min in 2017 to 2-min in 2018-2019. At each time step, two images are taken with different exposition times (1/100 sec and 1/2000 sec). From the initial $768 \times 1024$ pixel resolution, sky images are cropped and downscaled to a $128 \times 128$ pixel resolution (\citet{fengConvolutionalNeuralNetworks2022}). In addition, images are unwrapped on a rectangular regular grid (see Figure~\ref{fig:sky_images}) assuming median cloud heights (\citet{palettaTemporallyConsistentImagebased2020}) to limit the impact of the distortion on the scene representation and improve longer-term predictions (\citet{julianPreciseForecastingSky2021}).

\subsection{Satellite images}
\label{satellite_images}

\begin{figure*}%[H]%[ht!]%[h!] 
\centering
\begin{minipage}[b]{0.24\textwidth}
    \includegraphics[width=1.0\textwidth]{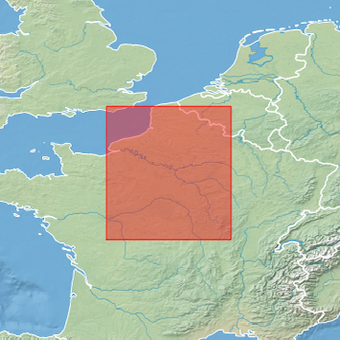}
  \end{minipage}
  \begin{minipage}[b]{0.24\textwidth}
    \includegraphics[decodearray={0.0 2.8}, width=1\textwidth]{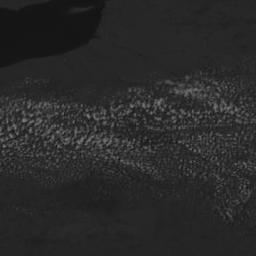}
  \end{minipage}
   \begin{minipage}[b]{0.24\textwidth}
    \includegraphics[decodearray={0.0 2.8 0.0 2.8 0.0 2.8}, width=1\textwidth]{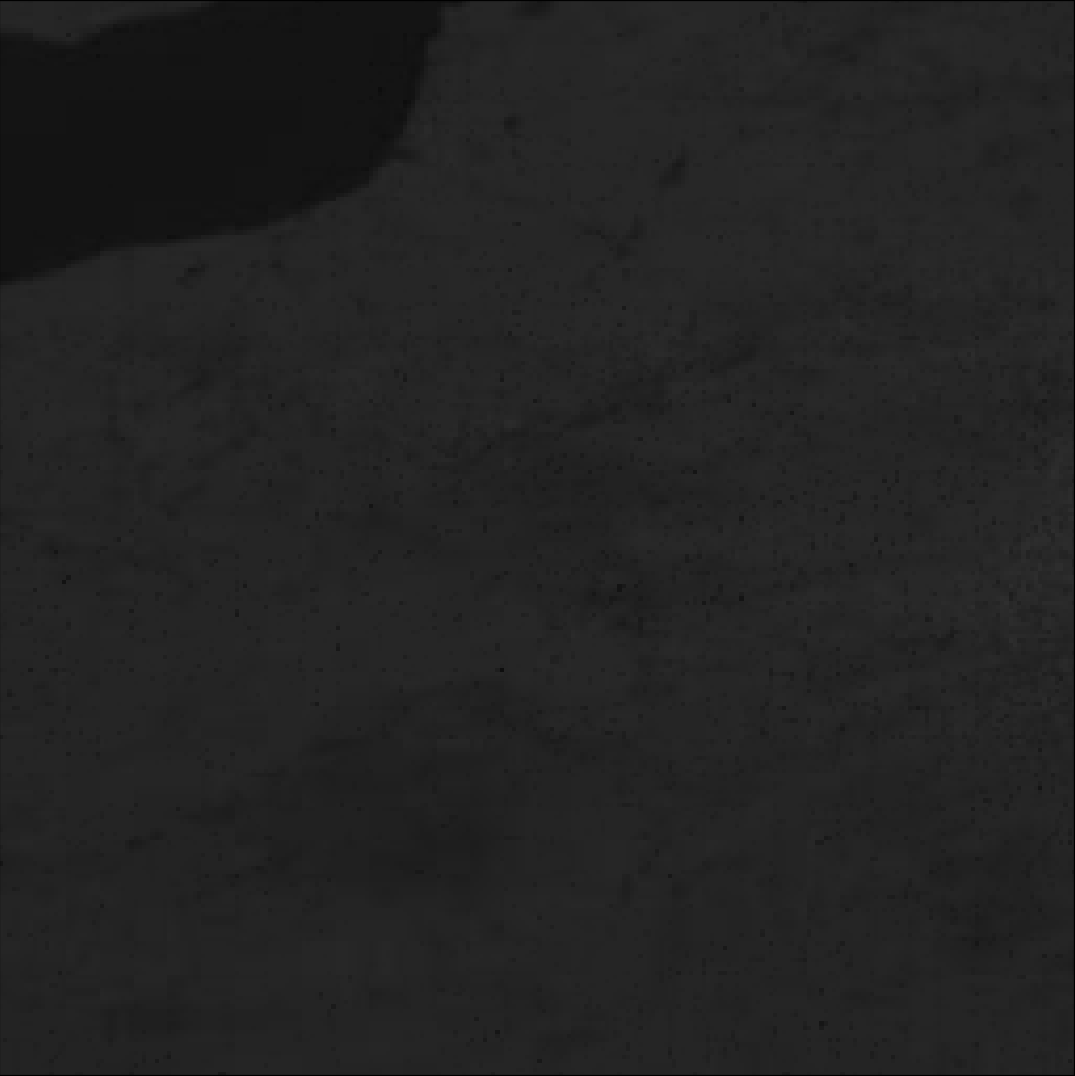}
  \end{minipage} 
  \begin{minipage}[b]{0.24\textwidth}
    \includegraphics[decodearray={0 1.8}, width=1\textwidth]{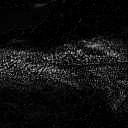}
    \end{minipage}
%  \begin{minipage}[b]{0.19\textwidth}
%    \includegraphics[width=1\textwidth]{Figures/20180510143000_01_sky_image_long_exposure.png}
%    \end{minipage}
%\vspace{-1.2\baselineskip}
\caption{From left to right: 1. Region centered on Sirta's laboratory covered by satellite observations, 2. Raw satellite image, 3. Ground albedo, 4. Cloud index map.}
\label{fig:satellite}
\end{figure*}

Satellite images originate from~\citet{eumetsatorganizationEUMETSATEuropeanOrganisation1991}~\footnote{Meteosat SEVIRI Rapid Scan image data : https://navigator.eumetsat.int/product/EO:EUM:DAT:MSG:MSG15-RSS}. The region covered by each frame corresponds to a $2.2^{\circ} \text{(latitude)}\times2.2^{\circ}$ (longitude) area centred on SIRTA's laboratory (48.713° N, 2.208° E) (see Figure~\ref{fig:satellite}). The resulting observation is a $256 \times 256$ pixels image downscaled by a factor of 4 to the same resolution as sky images ($128 \times 128$). Images are low pass filtered prior to down-sampling to prevent aliasing effects (\citet{parmarBuggyResizingLibraries2021}). The temporal resolution of satellite observations is 5-min.

\paragraph{Cloud index map}

Removing the background (sea and land) from satellite images is a traditional preprocessing step in satellite imagery (\citet{muellerRoleEffectiveCloud2011}). The resulting effective cloud albedo map measured in terms of cloud index (0: no cloud, 1: thick cloud) can be exploited to estimate the surface solar irradiance from cloud physical properties. DL models trained on cloud index maps try to focus more on the cloud dynamics, which constitutes the main source of short-term variability (\citet{mccandlessExaminingPotentialRandom2020, nielsenIrradianceNetSpatiotemporalDeep2021a}).

\begin{equation}
   \text{cloud index} (i, j, t) = \frac{p(i, j, t)- p_{min}(i, j, t-N:t)}{p_{max}(t) - p_{min}(i, j, t-N:t)}
   \label{equ:cloud_index}
\end{equation}

As shown in Equation~\ref{equ:cloud_index}, the cloud index corresponding to a given pixel ($i,j$) in the satellite image at time $t$ is computed from the current value $p(i, j, t)$, the minimum value of that pixel over the past $N$ days ($N=10$ here) at the same time $t$ ($p_{min}(i, j, t-N:t)$) and the maximum pixel value in the current image ($p_{max}(t)$) (\citet{palettaSPIN2021}). $p_{min}(i, j, t-N:t)$ and $p_{max}(t)$ correspond to the albedo of the ground and of thick clouds, respectively.

\subsection{Image transformation}

\begin{figure}%[H]%[ht!]%[h!] 
\centering
\begin{minipage}[b]{0.23\textwidth}
    \includegraphics[decodearray={0.0 1.8}, width=1.0\textwidth]{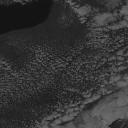}
  \end{minipage}
  \begin{minipage}[b]{0.23\textwidth}
    \includegraphics[decodearray={0.0 1.8}, width=1\textwidth]{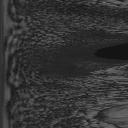}
  \end{minipage}
 \begin{minipage}[b]{0.23\textwidth}
    \includegraphics[decodearray={0.1 1.8}, width=1.0\textwidth]{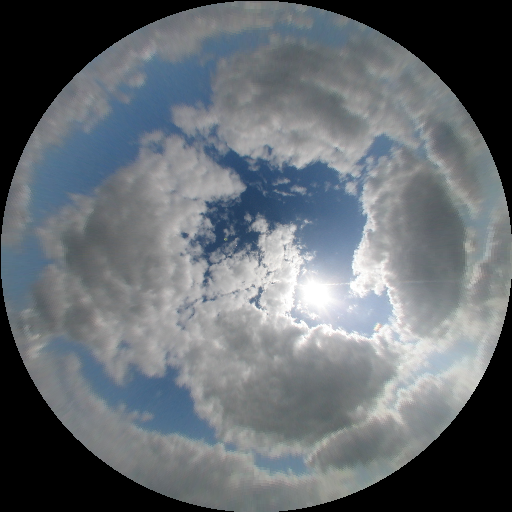}
  \end{minipage}
  \begin{minipage}[b]{0.23\textwidth}
    \includegraphics[width=1\textwidth]{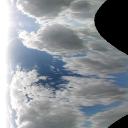}
  \end{minipage}
%\vspace{-1.2\baselineskip}
\caption{Satellite (Top) and sky images (Bottom). Original images are on the left and SPIN representation on the right. Notice the distortion caused by the use of polar coordinates in the right hand-side images. The resulting magnification of the area around the point of interest (sun or solar farm) leads to improved short-term predictions.}
\label{fig:spin}
\end{figure}

In addition to a recent focus on DL architectures for computer vision (CNN, LSTM, ConvLSTM, 3D-Convolutions, PhyDNet, Transformers, etc.) (\citet{Siddiqui2019, kongHybridApproachesBased2020, leguenDeepPhysicalModel2020a, palettaECLIPSEEnvisioningCloud2021}), data-centric approaches have been shown to significantly benefit image and video analysis. In particular, the SPIN method exploits the polar invariance of the irradiance forecasting task by representing the scene with polar coordinates centered on the sun in the sky image or the point of interest (e.g., a solar farm) in the satellite images (\citet{palettaSPIN2021}). The magnification of the area close to the centre of the polar coordinate system also benefits shorter-term predictions. Shift invariant convolutional architectures trained on this type of scene representation provide more accurate prediction and lower temporal lag, while training four times faster than methods which use rotations on the original image. Images resulting from applying the SPIN method to a satellite cloud index map and on an undistorted sky image are depicted in Figure~\ref{fig:spin}.

\section{Methodology}
\label{methodology}

\subsection{Deep learning architecture}
\label{architecture}

\begin{figure*}%[H]%[ht!]%[h!] 
\centering
\includegraphics[width=1.0\textwidth]{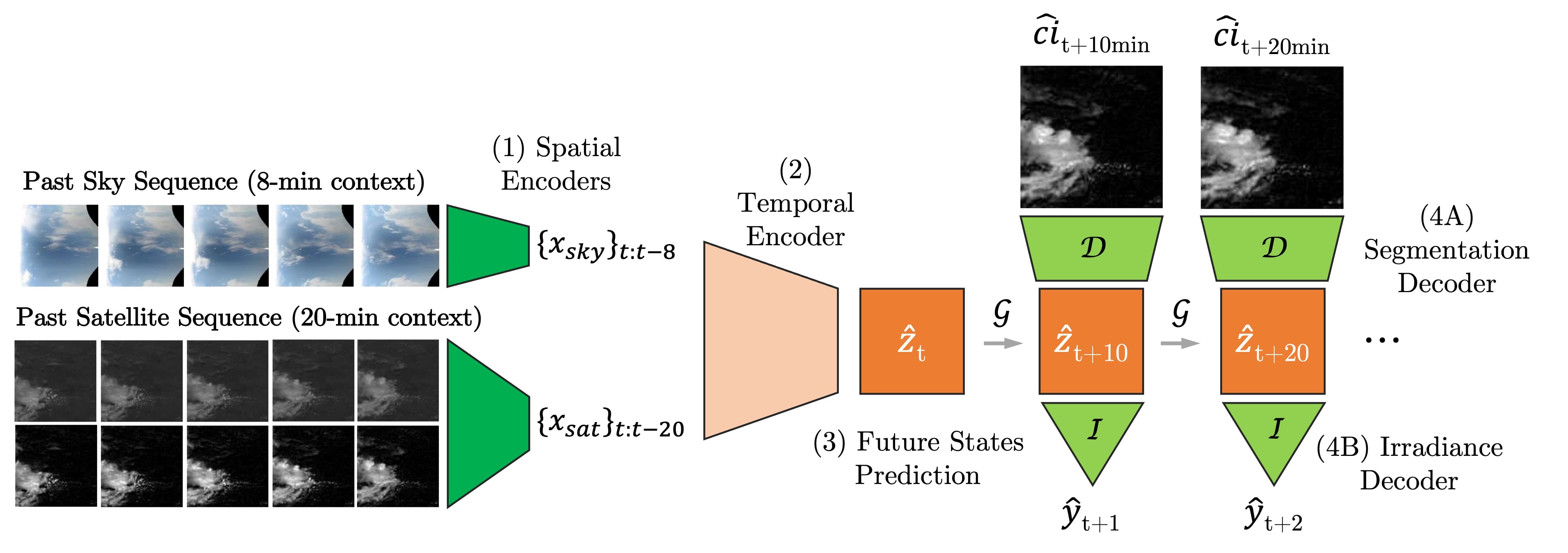}
%\vspace{-1.2\baselineskip}
\caption{DL architecture predicting future cloud index maps and corresponding irradiance levels from past sky and satellite images. Parallel encoders are used to extract spatial features from the sequences of past sky and satellite observations (raw images and cloud index maps). Future states are iteratively predicted from the output of the temporal encoder. Each state can then be decoded into different representations (cloud index maps and irradiance levels in the figure).}
\label{fig:model_architecture}
\end{figure*}

The backbone of the proposed DL architecture is ECLIPSE (\citet{palettaECLIPSEEnvisioningCloud2021}). It was shown to outperform other DL architectures on the irradiance forecasting task from both sky images and satellite images. The original model is composed of a spatial encoder (1) and a temporal encoder (2) to learn a spatio-temporal representation of past frames with 2D and 3D convolutions (Figure~\ref{fig:model_architecture}). Future states $z_{t}$ are then iteratively predicted through a Gated Recurrent Unit (GRU) module (3). Each state can then be decoded into a range of representations such as segmented image, RGB image, effective cloud albedo (4A), irradiance level (4B) or irradiance distribution.

In this study, we try to evaluate the benefit on short-term solar forecasts up to 1h of predicting from both sky and satellite images. The historical context of each sample consists of the five past sky images (2-min temporal resolution) and/or five past satellite images (5-min temporal resolution). These 2D inputs can be augmented with an extra irradiance channel (IC) encoding concomitant past normalised irradiance levels measured by the pyranometer on-site. The model predicts the 6 future states at a 10-min temporal resolution (10, 20, 30, 40, 50 and 60-min ahead). Each state is decoded into future satellite cloud index map $\hat{ci}$ and corresponding irradiance level $\hat{y}$ or irradiance distribution (classification over 100 same-size bins covering the range of irradiance levels). The overall loss function is a weighted sum of individual loss functions (irradiance: mean square error, irradiance distribution: cross-entropy loss function, cloud index: mean absolute error). 

Because sky and satellite observations are not spatially aligned, we feed each sequence into parallel spatial encoders. The resulting spatial representation is concatenated along the channel dimension before passing through the temporal encoder.

\subsection{Irregularly-sampled data}
\label{irregular_sampling}

The data used in this study originates from different sources detailed above. They were generated with different temporal resolutions (1-min for irradiance measurements, 1 to 2-min for sky images and 5-min for satellite images). Only considering samples whose most recent frame of each sequence type (sky and satellite images) were temporally concomitant lead to a small training set, thus poor performances. For this reason, samples whose last satellite image was taken less then 5-min prior to the sky image are also included in the training set.

The resulting batch of available samples was divided into three sets: training set (140k samples from 2017-2018), validation set (30k samples from even days of 2019) and test set (30k samples from odd days of 2019). The resulting distribution of samples by solar zenith angle, month and irradiance level is presented in~\ref{section:dataset_balance}. In addition, to investigate underlying structural effects in the overall forecast performance resulting from a particular distribution of weather conditions in the data, a set of 5 clear-sky days, 5 broken-sky days and 5 overcast days are selected from the test set for further quantitative and qualitative analyses. 

\subsection{Metrics}
\label{metrics}

\textbf{Forecast skill} Comparing absolute forecasting performances of models trained on different datasets is a hazardous task because of the range of aspects that can impact the magnitude of the average prediction error: weather conditions, distribution of clear-sky irradiance, daily time window considered for evaluation, etc. To overcome these limitations, a widely established method is to assess a model performance relative to a baseline model (\citet{yangVerificationDeterministicSolar2020a}). The resulting indicator called \textit{forecast skill} (FS), ranges from $-\infty$ to 100\% (Equation~\ref{equ:FS}). A FS higher than 0 indicates an improvement over the baseline, the closer to 100 the better.

\begin{equation}
  \text{Forecast Skill} = \left(1-\frac{\text{Error}_{model}}{\text{Error}_{baseline}}\right) \times 100
   \label{equ:FS}
\end{equation}

A recommended baseline model is the smart persistence model (SPM). Where the simple persistence model assumes no irradiance change over the given forecast horizon $\Delta t$ (Equation~\ref{equ:persitence}), the SPM takes into account the diurnal changes of the extra-terrestrial irradiance $y_{clr}$ over the forecast horizon $\Delta t$ (Equation~\ref{equ:smart_persistence}). The longer the horizon, the higher the impact of the diurnal parameter on the error. This out-of-the-atmosphere irradiance $y_{clr}$ is modelled by a clear-sky model called McClear (\citet{lefevreMcClearNewModel2013}).

\begin{equation}
   \hat{y}(t+\Delta t) = y(t)
   \label{equ:persitence}
\end{equation}

\begin{equation}
   \hat{y}(t+\Delta T) = \frac{y_{clr}(t+\Delta t)}{y_{clr}(t)} \,  y(t)
   \label{equ:smart_persistence}
\end{equation}

\vspace{0.5\baselineskip}

\textbf{Critical event prediction} A limitation of metrics such as the FS based on average errors such as the mean absolute error or the root mean square error (RMSE), is a strong dependency on the overall error distribution: the performance of a model on sharp irradiance changes corresponding to rare but critical events, might be over-weighted by more common small changes in the aggregated sum of errors. For this reason, we also evaluate the model on five clear-sky and five overcast days (low variability) in addition to five broken-sky days (high variability) from the test set. Furthermore, we report the 95\% quantile of the largest errors in the test set.

\vspace{0.5\baselineskip}

\textbf{Continuous ranked probabilistic score (CRPS)} The CRPS is a scoring function enabling objective probabilistic forecast evaluation (\citet{lauretVerificationSolarIrradiance2019, hersbachDecompositionContinuousRanked2000}) and cross-site comparison.
The CRPS measures the difference between cumulative distribution functions $F$ as shown in Equation~\ref{equ:crps}. In irradiance forecasting, the ground truth $F_{target}$ is the cumulative  distribution of the step function that jumps from 0 to 1 at the value of the measured irradiance level. $F_{model}$ is the cumulative distribution of the irradiance as predicted by the model. In Equation~\ref{equ:crps}, the CRPS is computed as an average over a set of N predictions:

\begin{equation}
   CRPS = \frac{1}{N} \sum_{n=1}^{N} \int_{-\infty}^{+\infty} [F_{target}^{n}(x)-F_{model}^{n}(x)]^2 dx
   \label{equ:crps}
\end{equation}

The CRPS is dimensionally the same as the variable being predicted (W/$\text{m}^2$ for the GHI) and it is equal to the mean absolute error if the prediction of the model is deterministic. Similarly to deterministic predictions, probabilistic performances can be expressed relative to the SPM using the FS score.

\begin{table*}[ht!]
%\vspace{-1\baselineskip}
\caption{Comparative study on the different types of image transformation: raw satellite image, close-up on the centre of the image corresponding to a fourth of the original spatial coverage, raw sky images and SPIN transform applied to all. The coefficient $\alpha$ corresponding to the image prediction loss $L_{image}$ in Equation~\ref{equ:total_loss_with_distribution} is set to $0$.}
\vspace{-0.5\baselineskip}
\begin{center}
\begin{tabular}{lcccccccc}
\hline
\noalign{\vskip 1mm}
 & & \multicolumn{3}{c}{RMSE $\downarrow$ [W/$\text{m}^2$] (Forecast Skill $\uparrow$ [\%])} & & \multicolumn{3}{c}{95\% Quantile $\downarrow$ [W/$\text{m}^2$]}\\

 Forecast Horizons & $\mid$ & 10-min & 30-min & 60-min & $\mid$ & 10-min & 30-min & 60-min \\
\hline\hline
\noalign{\vskip 2mm}
%Smart Pers. && 144.8 (0\%) & 166.0 (0\%) & 174.4 (0\%) & & 348.8 & 402.6 & 422.2 \\
Smart Pers. && 144.6 (0\%) & 165.1 (0\%) & 172.4 (0\%) & & 348.0 & 401.1 & 418.8 \\
\noalign{\vskip 1mm}
\multicolumn{3}{l}{Satellite images (+ Raw sky images):}   &  &  & &  &  &  \\
\noalign{\vskip 0.5mm}
 - Raw && 120.4 (16.7\%) & 133.0 (19.5\%) & 140.4 (18.5\%) & & 270.2 & 299.3 & 315.6 \\
\noalign{\vskip 0.3mm}
 - Close-up && 119.4 (17.4\%) & \textbf{131.6 (20.3\%)} & 139.0 (19.4\%) & & \textbf{268.6} & \textbf{297.4} & \textbf{311.5} \\
\noalign{\vskip 0.3mm}
 - Raw (SPIN) && 120.5 (16.6\%) & 132.5 (19.8\%) & \textbf{138.2 (19.8\%)} & & 275.6 & 303.4 & 314.5 \\
\noalign{\vskip 0.3mm}
 - Close-up (SPIN) && \textbf{119.2 (17.5\%)} & 132.8 (19.6\%) & 138.9 (19.4\%) & & 272.6 & 301.6 & 315.2 \\
\noalign{\vskip 1mm}

\noalign{\vskip 1mm}
\multicolumn{3}{l}{Satellite images (+ SPIN sky images):}   &  &  & &  &  &  \\
\noalign{\vskip 0.5mm}
 - Raw && 116.3 (19.5\%) & 129.2 (21.8\%) & \textbf{134.9 (21.8\%)} && \textbf{262.9} & \textbf{293.9} & 306.1 \\
\noalign{\vskip 0.3mm}
 - Close-up && \textbf{115.9 (19.9\%)} & 128.7 (22.1\%) & 136.0 (21.1\%) && 266.2 & 294.1 & \textbf{305.5} \\
\noalign{\vskip 0.3mm}
 - Raw (SPIN) && 199.0 (17.7\%) & 130.8 (20.8\%) & 135.4 (21.4\%) && 269.1 & 297.4 & 306.9 \\
\noalign{\vskip 0.3mm}
 - Close-up (SPIN) && 116.4 (19.5\%) & \textbf{126.7 (23.3\%)} & 136.9 (20.6\%) && 268.1 & 297.0 & 311.6 \\
\noalign{\vskip 1mm}

\hline
\end{tabular}
\end{center}
\label{tab:results_sat_images}
\end{table*}

\section{Experimental results}
\label{results}

We conducted a range of experiments to quantify the respective benefits of predicting solar irradiance from sky and satellite images. For each forecast horizon, the model training iteration corresponding to the best FS performance on the validation set is used to calculate the test performances for that horizon. All results presented below are averaged over two trainings with different random initialisations.

\subsection{Comparative study on input images}
\label{input_images}

Regarding the scene representation, there is a clear advantage for the proposed architecture to represent the sky image with polar coordinates (SPIN) compared to raw sky images as shown in~\citet{palettaSPIN2021}. The resulting FS increases by at least 1\% for all forecast horizons and all satellite image types (Table~\ref{tab:results_sat_images}). For satellite images, the best scene representations seem to be those close up on the image centre. The corresponding SPIN representation of the effective cloud albedo map gives similar FS scores (a 10 to 60-min average of 21.8\% for both), but its resulting 95\% error quantile is higher for all horizons.
As a result, the image representation setup considered in this study on intra-hour solar forecasting is a combination of the satellite image close-up and sky images in polar coordinates (SPIN).

\subsection{Loss weighting}
\label{losses}

\begin{figure}%[H]%[ht!]%[h!] 
\centering
\includegraphics[width=0.5\textwidth]{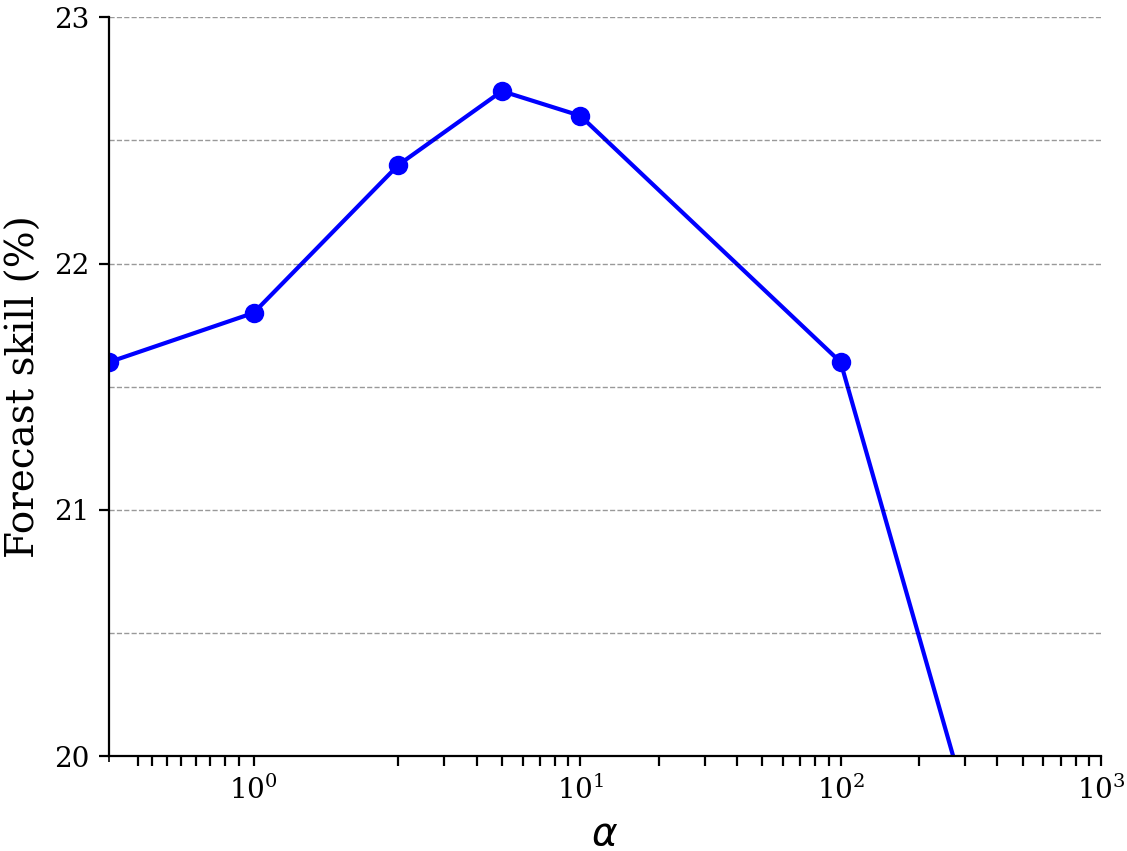}
%\vspace{-1.2\baselineskip}
\caption{FS averaged over all 6 forecast horizons (10 to 60-min ahead) for different values of $\alpha$. Image and irradiance loss weighting shows a global optimum ($\alpha \approx 5$ here) beyond which the overall performance drops.}
\label{fig:horizons_loss_weighting}
\end{figure}

\begin{figure}%[H]%[ht!]%[h!] 
\centering
\includegraphics[width=0.5\textwidth]{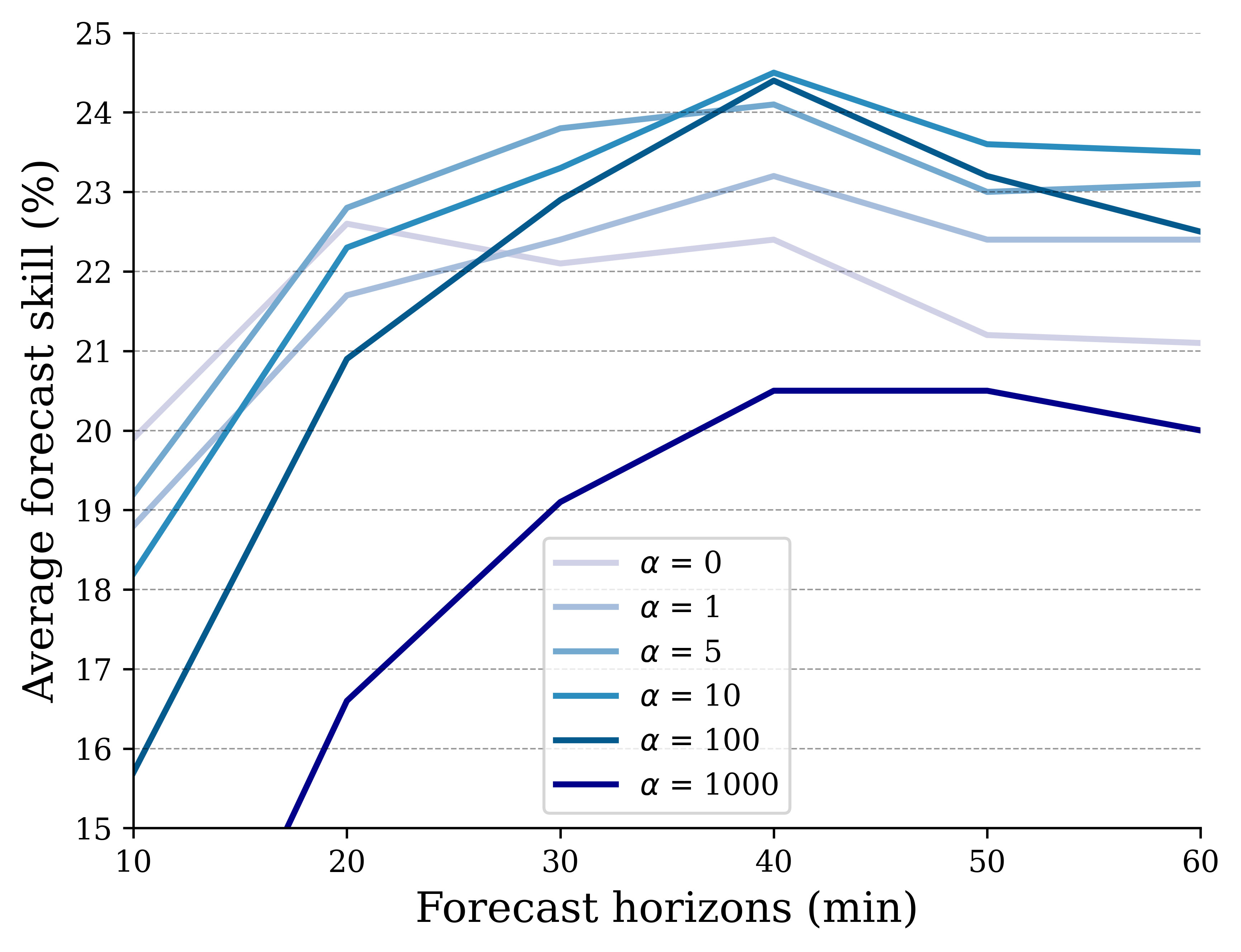}
%\vspace{-1.2\baselineskip}
\caption{FS over all 6 forecast horizons (10 to 60-min ahead) for different values of $\alpha$. Up to a certain value of $\alpha$, long-term irradiance predictions ($\geq$20-min) benefit from the supervision signal induced by predicting future images at the cost of higher shorter-term errors (10-min).}
\label{fig:horizons_weighting}
\end{figure}

The overall loss $L_{total}$ used to train the deterministic model is a weighted sum of the individual losses supervising the two competing objectives of the model: irradiance and image predictions (Equation~\ref{equ:total_loss_with_distribution}) (mean square error in both cases):

\begin{equation}
  L_{total} = L_{irradiance} + \alpha L_{image}
   \label{equ:total_loss_with_distribution}
\end{equation}

Here, we try to quantify the impact of the weighting parameter $\alpha$ on the irradiance prediction task. Figure~\ref{fig:horizons_loss_weighting} shows the value of the average FS (10 to 60-min lead times) for different values of $\alpha$. The FS appears to increase with $\alpha$ up to a certain point ($\alpha=5$ here). In other words, the irradiance forecasting task benefits from some co-supervision based on video prediction as suggested in~\citet{palettaECLIPSEEnvisioningCloud2021}. As expected, increasing the relative weight of $L_{image}$ in the total loss functions ($\alpha > 5$ here) leads to lower irradiance forecasting performances

Interestingly, the value of $\alpha$ impacts the accuracy of each forecast horizon differently (Figure~\ref{fig:horizons_weighting}). The higher the weight on $L_{irradiance}$, the better very short-term predictions (10-min ahead). Up to a certain value of $\alpha$, increasing the share of $L_{image}$ in $L_{total}$ leads to improved longer term-forecasts with an optimum value minimising the average error ($\alpha = 5$ here).

\begin{table*}[ht!]
\caption{Relative benefits of sky and satellite images in vision-based intra-hour irradiance forecasting. The model is trained to predict future satellite images and irradiance values from past sky images (SPIN) and satellite images (Close-up). The coefficient $\alpha$ corresponding to the image prediction loss $L_{image}$ in Equation~\ref{equ:total_loss_with_distribution} is set to $5$ (Figure~\ref{fig:horizons_loss_weighting}).}
\vspace{-0.8\baselineskip}
\begin{center}
\begin{tabular}{lcccccccc}
\hline
\noalign{\vskip 1mm}
 & & \multicolumn{3}{c}{RMSE $\downarrow$ [W/$\text{m}^2$] (Forecast Skill $\uparrow$ [\%])} & & \multicolumn{3}{c}{95\% Quantile $\downarrow$ [W/$\text{m}^2$]}\\
 Forecast Horizons & $\mid$ & 10-min & 30-min & 60-min & $\mid$ & 10-min & 30-min & 60-min \\
\hline\hline
\noalign{\vskip 2mm}
%Smart Pers. &&  (0\%) &  (0\%) &  (0\%) & &  &  &  \\
%\noalign{\vskip 1mm}
%ECLIPSE &&  &  &  & &  &  &  \\
%\noalign{\vskip 0.3mm}
 - Satellite Images && 138.9 (3.9\%) & 143.5 (13.1\%) & 149.2 (13.5\%) && 311.7 & 321.4 & 326.7 \\
 - Satellite Images (+IC) && 124.7 (13.7\%) & 135.6 (17.9\%) & 143.0 (17.0\%) && 289.0 & 308.1 & 317.5 \\
\noalign{\vskip 2mm}
 - Sky Images && 112.6 (22.1\%) & 129.7 (21.5\%) & 138.1 (19.9\%) && 253.3 & 289.2 & 314.5 \\
 - Sky Images (+IC) && \textbf{110.2 (23.8\%)} & 128.0 (22.5\%) & 136.3 (20.9\%) && \textbf{250.7} & 293.4 & 311.4 \\
\noalign{\vskip 2mm}
 - Sky and Satellite Images && 116.8 (19.2\%) & \textbf{125.8 (23.8\%)} & 132.6 (23.1\%) && 267.4 & \textbf{286.6} & 298.5 \\
 - Sky and Satellite Images (+IC) && 115.6 (20.0\%) & 125.9 (23.8\%) & \textbf{131.5 (23.8\%)} && 269.8 & 289.1 & \textbf{295.9} \\
\noalign{\vskip 1mm}
\hline
\end{tabular}
\end{center}
\label{tab:results_eclipse}
\end{table*}

\subsection{Ablation study}
\label{multiobjectives}

\begin{figure}%[H]%[ht!]%[h!] 
\centering
\includegraphics[width=0.5\textwidth]{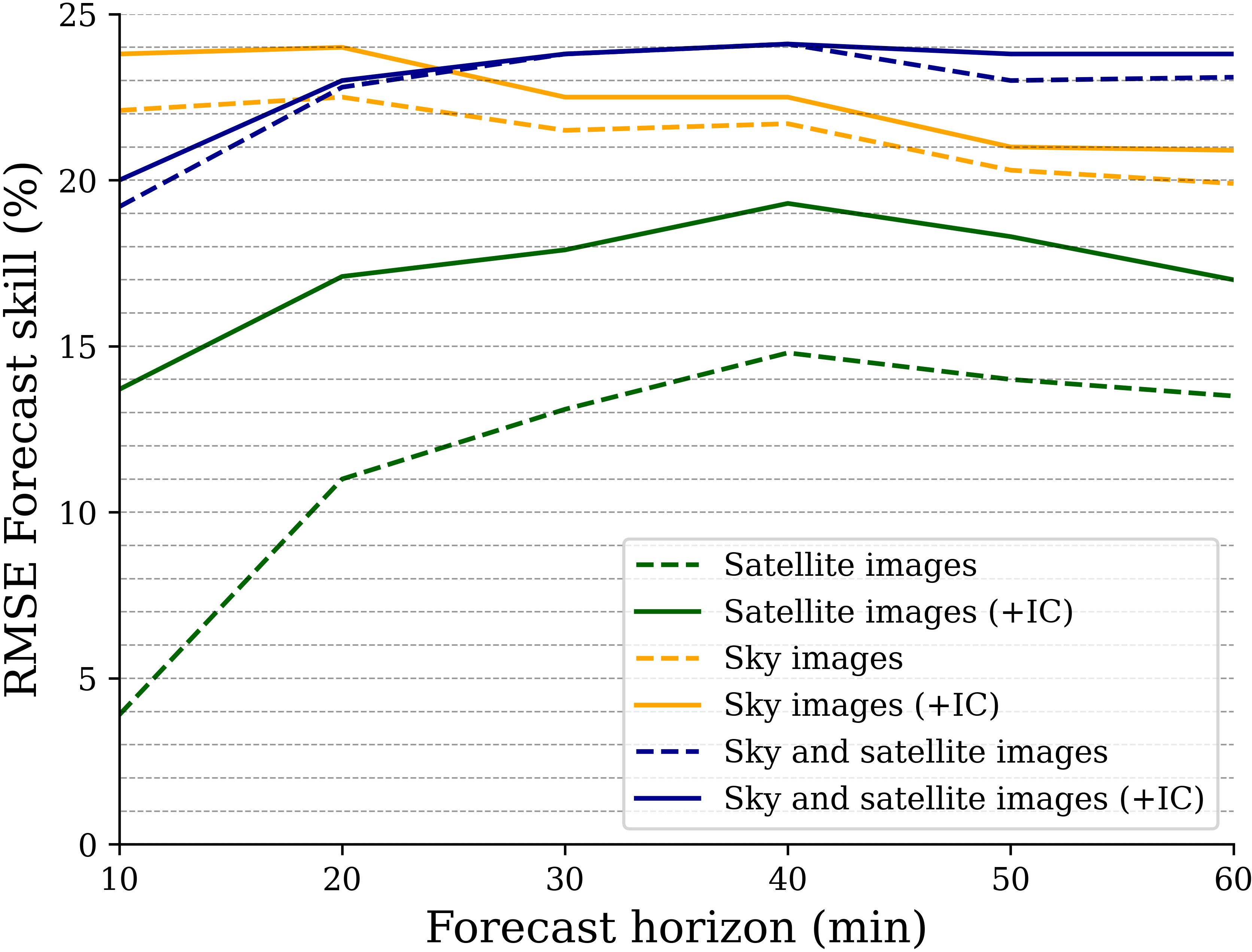}
%\vspace{-1.2\baselineskip}
\caption{Comparative performance of the different types of observation in intra-hour irradiance forecasting using the FS based on the RMSE metric. The gain of using satellite images in addition to sky images is visible from a 25-min lead time. Models trained on satellite observations benefit the most from the additional irradiance channel (IC)}
\label{fig:horizons_fs}
\end{figure}

In previous works, sky and satellite observations have been used separately for different forecast windows: up to 20-30min for sky images and from 15-min for satellite images. Here, we try to evaluate the benefit of combining both data sources given the prediction horizon.

Figures~\ref{fig:horizons_fs} and~\ref{fig:horizons_95quantile} show that for longer-term forecasts (25 to 60-min ahead), relying on both data types improves the average performance and high irradiance change prediction. In particular, the resulting FS increases by about 10\% over models using satellite images only (Table~\ref{tab:results_eclipse}). In comparison, the hybrid model (sky and satellite images) increases its FS by 2-3\% only compared to models trained to forecast solar irradiance from past sky images alone. For short-term forecasting (10-min ahead) however, the hybrid model is worse in both FS and high irradiance change prediction. This might be caused by a bottleneck issue: the neural network benefits more from focusing on the most valuable source of information (sky images in short-term forecasting), rather than allocating a share of its parameters to extracting features from a less informative source (satellite images which have a lower temporal resolution). In addition, the cloud index video prediction task being significantly harder when no input past cloud maps are fed to the model, the resulting supervision signal weakens, hence the training resembles the one based on irradiance predictions alone ($\alpha=0$). As shown in Figure~\ref{fig:horizons_weighting}, this configuration benefits shorter-term forecasts more (about 1\% FS difference). Overall, Figure~\ref{fig:horizons_fs} highlights the advantage of combining both observations from a 25-min lead time.

Moreover, adding an extra irradiance channel (IC) improves performances in almost all configurations, the most significant gain being for models trained on satellite observations (Figure~\ref{fig:horizons_fs}). This highlights the difficulty for DL models to correlate an image with the corresponding local irradiance level (\citet{palettaECLIPSEEnvisioningCloud2021}). Adding this extra information through the IC makes up for some of the performance gap with models relying on sky images.

\subsection{High irradiance shift forecasting}
\label{event_prediction}

\begin{figure}%[H]%[ht!]%[h!] 
\centering
\includegraphics[width=0.5\textwidth]{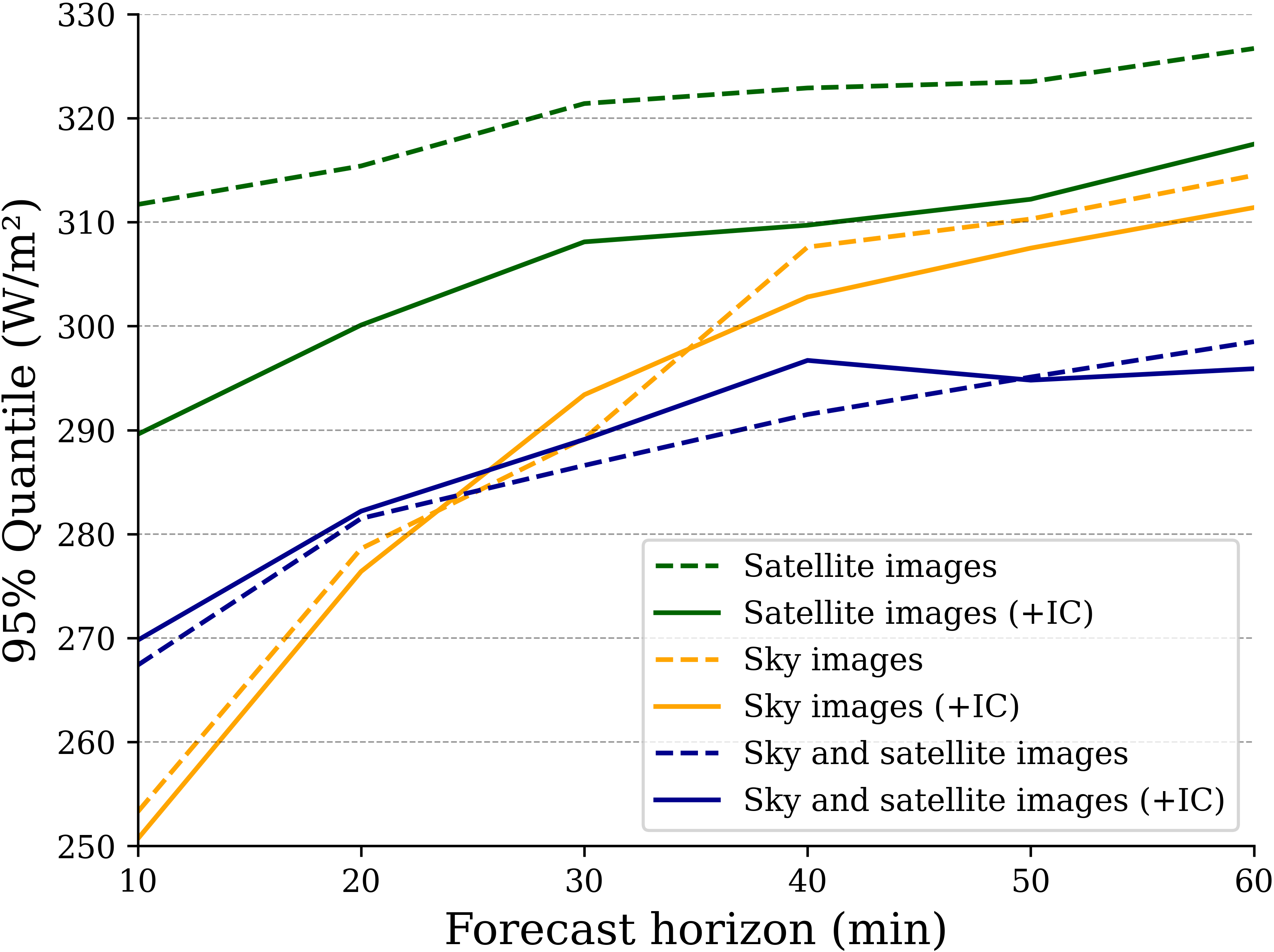}
%\vspace{-1.2\baselineskip}
\caption{Comparative performance of the different types of observation in large irradiance shift prediction using the 95\% quantile. The gain of using satellite images in addition to sky images appears from a 25-min lead time. Models trained on satellite observations benefit the most from the additional irradiance channel (IC)}
\label{fig:horizons_95quantile}
\end{figure}

Predicting large irradiance changes is critical in many solar applications (hybrid power plants, energy trading, etc.). To isolate these critical events, the 95\% quantile of the sorted forecast errors for different input configurations is reported in Table~\ref{tab:results_eclipse}. Most of the observations mentioned in the previous section still hold. In particular, the model trained on sky images outperforms those using satellite images on very short-term predictions (10-min lead time). In contrast, the hybrid configuration (satellite and sky images) is the most accurate on 30 to 60-min ahead forecasts (Figure~\ref{fig:horizons_95quantile}).  Although the benefit of the extra IC is clear for the satellite-based model, it is not for the other two configurations. Specifically, the hybrid model slightly underperforms in 10 to 40-min ahead forecasts when conditioned on past irradiance observations.

\subsection{Sky conditions}
\label{sky_conditions}

\begin{table*}[ht!]
%\vspace{-1\baselineskip}
\caption{Comparative study on different weather conditions: clear sky, broken-sky and overcast. The input data are the close-up on the satellite image (SO), the SPIN transform of the sky image (SI) and the irradiance channel (IC). The $\alpha$ coefficient corresponding to the image prediction loss $L_{image}$ in Equation~\ref{equ:total_loss_with_distribution} is set to $5$ (Figure~\ref{fig:horizons_loss_weighting}).}
\vspace{-0.8\baselineskip}
\begin{center}
\begin{tabular}{cccccccccc}
\hline
\noalign{\vskip 1mm}
 & & & & \multicolumn{6}{c}{RMSE $\downarrow$ [W/$\text{m}^2$] (Forecast Skill $\uparrow$ [\%])} \\

 \multicolumn{3}{c}{Forecast horions} & $\mid$ & 10-min & 20-min & 30-min & 40-min & 50-min & 60-min \\
\hline\hline
%\noalign{\vskip 2mm}
%Smart Pers. && 144.8 (0\%) & 166.0 (0\%) & 174.4 (0\%) & & 348.8 & 402.6 & 422.2 \\
%Smart Pers. && 144.6 (0\%) & 165.1 (0\%) & 172.4 (0\%) & 348.0 & 401.1 & 418.8 \\
\noalign{\vskip 1mm}
%\multicolumn{3}{c}{Clear-sky} & & &  &  & & &  \\
SO & SI & IC && \multicolumn{6}{c}{Clear-sky}  \\
\noalign{\vskip 0.3mm}
\checkmark & & && 50.2 (-283\%) & 48.2 (-272\%) & 46.8 (-244\%) & 47.1 (-236\%) & 51.0 (-261\%) & 59.0 (-312\%) \\
\noalign{\vskip 0.3mm}
\checkmark & & \checkmark && 26.2 (-100\%) & 33.6 (-159\%) & 38.1 (-180\%) & 43.0 (-207\%) & 48.2 (-242\%) & 54.5 (-281\%) \\
\noalign{\vskip 0.3mm}
  & \checkmark & && 26.5 (-102\%) & 29.6 (-128\%) & 30.5 (-125\%) & 28.0 (-100\%) & 27.1 (-92\%) & \textbf{31.7 (-122\%)} \\
\noalign{\vskip 0.3mm}
 & \checkmark & \checkmark && 24.1 (-84\%) & 26.6 (-105\%) & 27.6 (-103\%) & 27.3 (-95\%) & 30.4 (-115\%) & 38.0 (-166\%) \\
\noalign{\vskip 0.3mm}
\checkmark & \checkmark & && 28.9 (-121\%) & 37.8 (-192\%) & 35.3 (-160\%) & 32.5 (-132\%) & 33.3 (-136\%) & 40.8 (-186\%) \\
 \noalign{\vskip 0.3mm}
\checkmark & \checkmark & \checkmark && \textbf{21.2 (-62\%)} & \textbf{21.8 (-68\%)} & \textbf{19.9 (-46\%)} & \textbf{20.8 (-49\%)} & \textbf{26.4 (-87\%)} & 35.7 (-150\%) \\

\noalign{\vskip 3mm}

%\multicolumn{3}{c}{Broken-sky} & & &  &  & & &  \\
SO & SI & IC && \multicolumn{6}{c}{Broken-sky} \\
\checkmark  & & && 179.4 (19\%) & 178.4 (22\%) & 179.1 (24\%) & 180.2 (26\%) & 182.6 (25\%) & 189.1 (23\%) \\
\noalign{\vskip 0.3mm}
\checkmark &  & \checkmark && 175.3 (21\%) & 181.2 (21\%) & 182.5 (22\%) & 190.2 (22\%) & 193.8 (20\%) & 196.9 (20\%) \\
\noalign{\vskip 0.3mm}
 & \checkmark & && 162.4 (27\%) & 167.0 (27\%) & 171.0 (27\%) & 174.2 (29\%) & \textbf{175.7 (28\%)} & \textbf{177.0 (28\%)} \\
\noalign{\vskip 0.3mm}
& \checkmark & \checkmark && \textbf{161.0 (28\%)} & \textbf{164.9 (28\%)} & \textbf{169.7 (28\%)} & \textbf{173.0 (29\%)} & 178 (27\%) & 182.5 (26\%) \\
\noalign{\vskip 0.3mm}
\checkmark & \checkmark & && 164.7 (26\%) & 170.3 (25\%) & 172.5 (26\%) & 174.6 (28\%) & 178.0 (27\%) & 181.6 (26\%) \\
 \noalign{\vskip 0.3mm}
\checkmark & \checkmark & \checkmark && 164.1 (26\%) & 167.8 (26\%) & 171.3 (27\%) & 174.0 (29\%) & 176.7 (27\%) & 179.7 (27\%) \\

\noalign{\vskip 3mm}

%\multicolumn{3}{c}{Overcast} & & &  &  & & &  \\
SO & SI & IC && \multicolumn{6}{c}{Overcast} \\
\checkmark & & && 122.4 (-138\%) & 125.1 (-95\%) & 128.9 (-101\%) & 131.8 (-108\%) & 133.7 (-100\%) & 135.1 (-98\%) \\
 \noalign{\vskip 0.3mm}
\checkmark & & \checkmark && 67.7 (-31\%) & 75.4 (-18\%) & 77.4 (-21\%) & 80.3 (-27\%) & 82.7 (-23\%) & 82.5 (-21\%) \\
\noalign{\vskip 0.3mm}
  & \checkmark & && 53.6 (-4\%) & 62.4 (2\%) & 64.2 (0\%) & 66.3 (-5\%) & 70.2 (-5\%) & 69.8 (-2\%) \\
\noalign{\vskip 0.3mm}
& \checkmark & \checkmark && \textbf{49.8 (4\%)} & \textbf{59.7 (7\%)} & \textbf{60.1 (6\%)} & \textbf{60.0 (5\%)} & \textbf{63.3 (6\%)} & \textbf{65.3 (4\%)} \\
\noalign{\vskip 0.3mm}
\checkmark & \checkmark & && 60.7 (-17\%) & 65.5 (-2\%) & 64.2 (0\%) & 65.3 (-3\%) & 68.3 (-2\%) & 69.0 (-1\%) \\
\noalign{\vskip 0.3mm}
\checkmark & \checkmark & \checkmark && 65.3 (-26\%) & 79.5 (-24\%) & 80.8 (-26\%) & 81.9 (-29\%) & 82.3 (-23\%) & 82.5 (-21\%) \\

\noalign{\vskip 1mm}
\hline
\end{tabular}
\end{center}
\label{tab:results_sky_conditions_all}
\end{table*}

\begin{figure}%[H]%[ht!]%[h!] 
\centering
\includegraphics[width=0.48\textwidth]{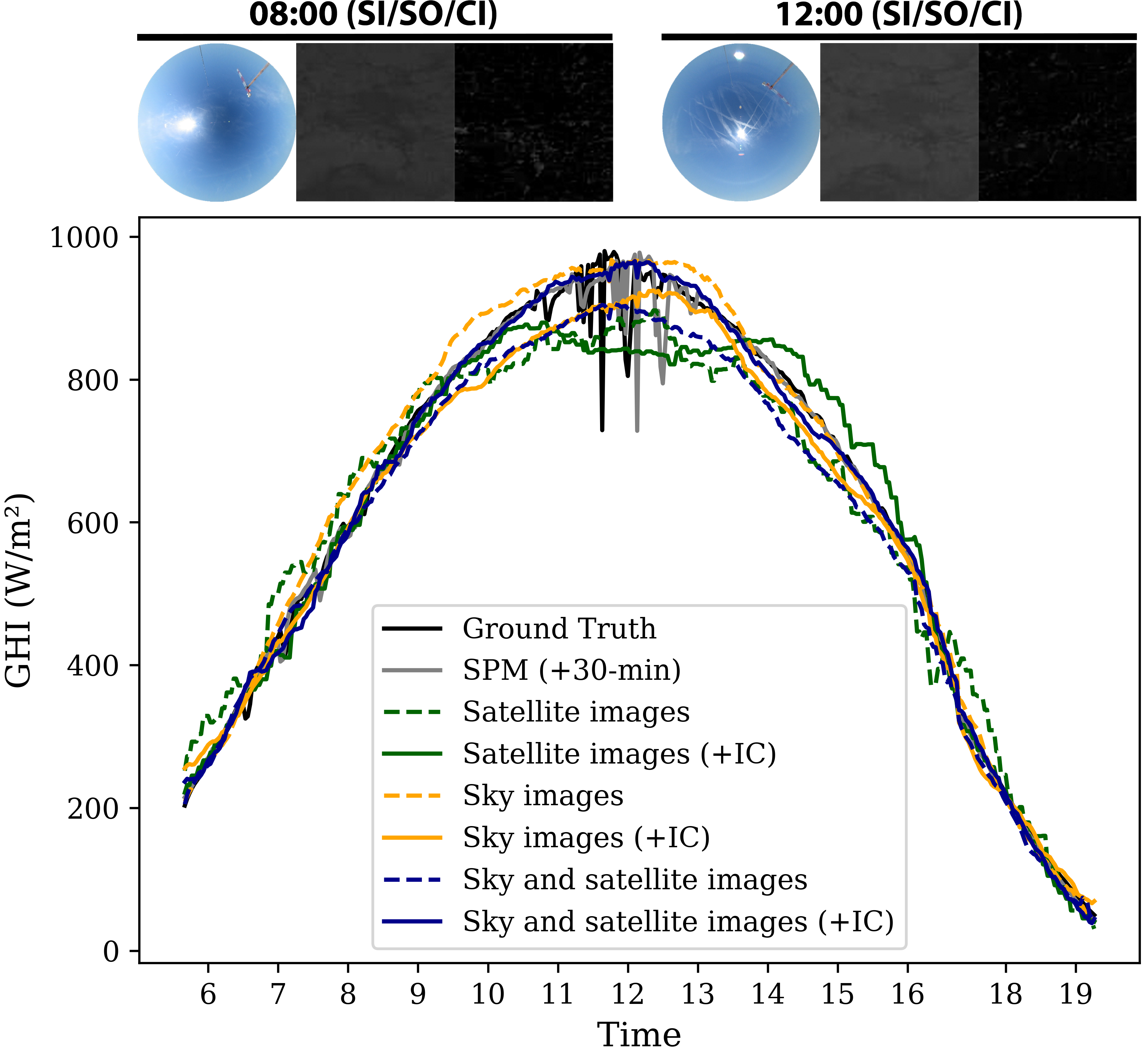}
%\vspace{-1.2\baselineskip}
\caption{30-min ahead prediction curves over a clear-sky day (05/07/2019). The largest forecast errors happen in the middle of the day when the clear-sky irradiance is the highest. The irradiance channel benefits the most to the model trained on both sky and satellite images. Notice that the short downward peaks visible around noon are caused by contrails. SI: Sky image, SO: Satellite observation, CI: Cloud index.}
\label{fig:clear_sky_day}
\end{figure}

\begin{figure}%[H]%[ht!]%[h!] 
\centering
\includegraphics[width=0.48\textwidth]{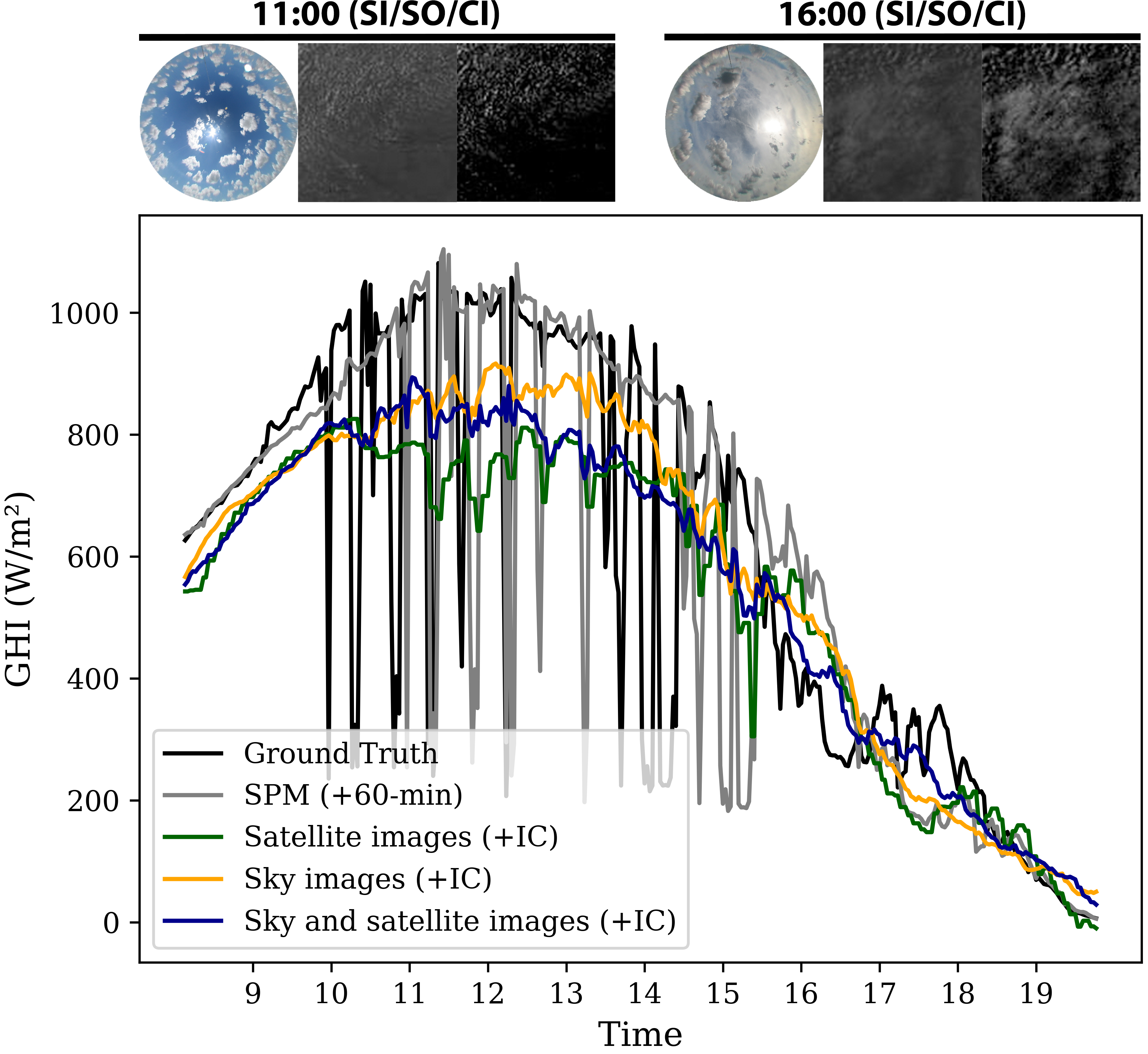}
%\vspace{-1.2\baselineskip}
\caption{60-min ahead prediction curves over a broken-sky day (17/06/2019). All model forecasts follow a similar trend by predicting intermediate irradiance levels to minimise the risk of large errors as shown in~\citet{palettaBenchmarkingDeepLearning2021c}. SI: Sky image, SO: Satellite observation, CI: Cloud index.}
\label{fig:broken_sky_day}
\end{figure}

\begin{figure}%[H]%[ht!]%[h!] 
\centering
\includegraphics[width=0.48\textwidth]{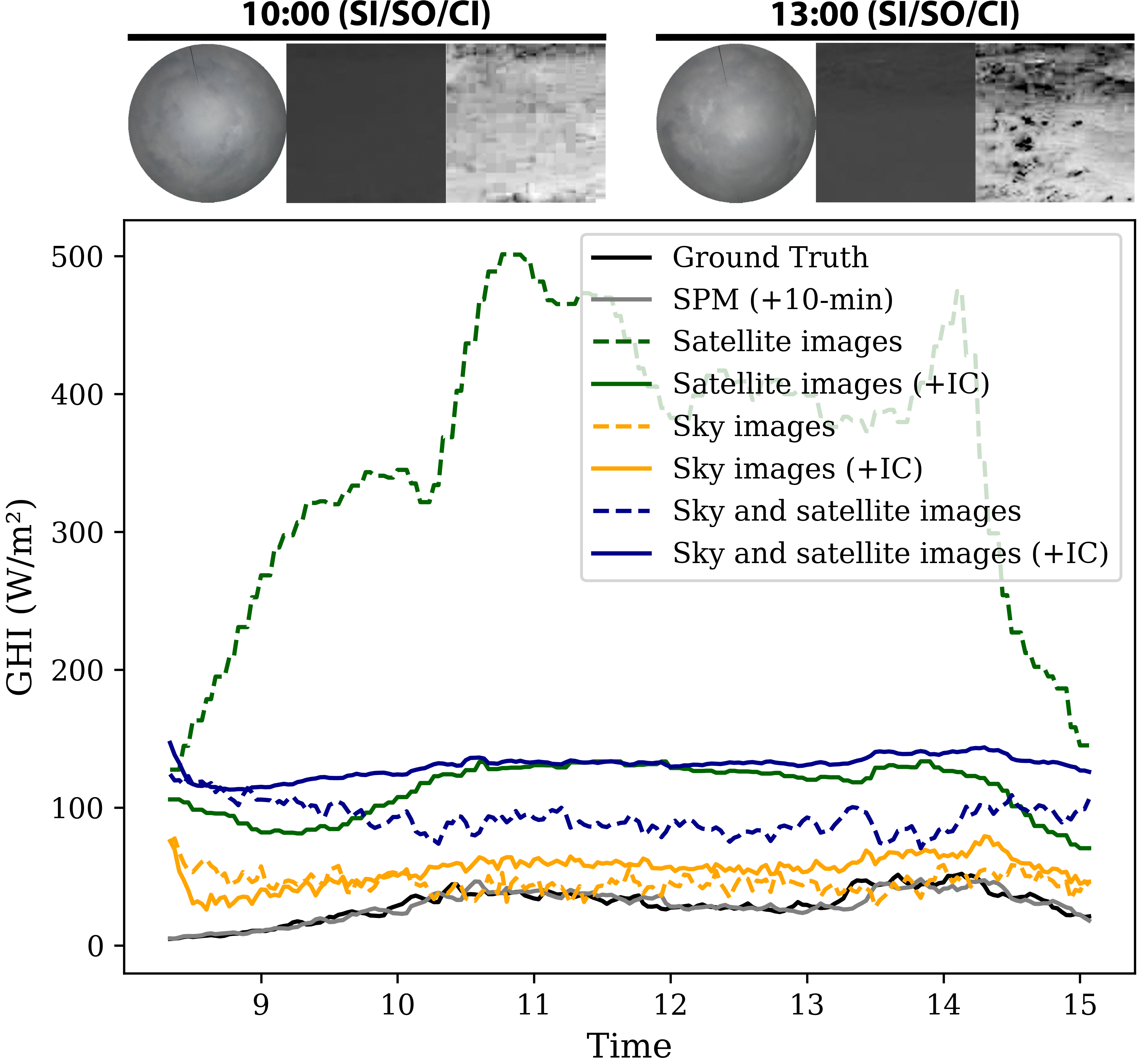}
%\vspace{-1.2\baselineskip}
\caption{10-min ahead prediction curves over an overcast day (01/01/2019). The additional irradiance channel strongly benefits the model trained on satellite images only. SI: Sky image, SO: Satellite observation, CI: Cloud index.}
\label{fig:overcast_day}
\end{figure}

\begin{figure}%[H]%[ht!]%[h!] 
\centering
\includegraphics[width=0.48\textwidth]{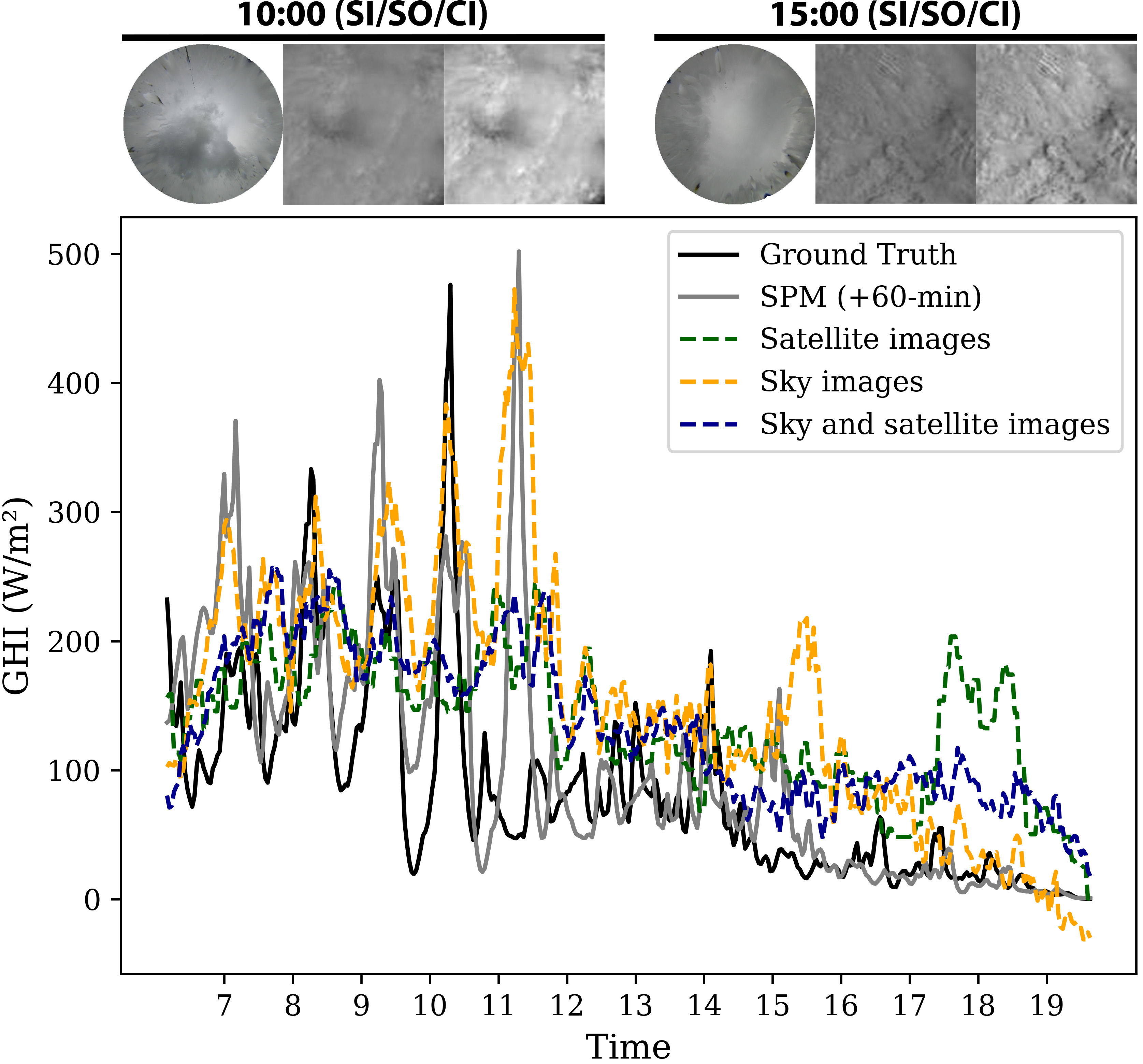}
%\vspace{-1.2\baselineskip}
\caption{60-min ahead prediction curves over an overcast day (05/06/2019). The model trained on sky images suffers from a heavy inertia resulting in predictions closely following the SPM during high irradiance events (7:00, 9:30, 11:20).}
\label{fig:overcast_day_inertia}
\end{figure}

The overall performance of a model averaged over a large number of days hides the specificity of weather dependent performances. For instance, computer vision approaches are expected to perform the best in cloudy conditions, whereas statistical models benefit from highly auto-correlated irradiance levels in cloud-free days. In practice, the cumulative prediction error of a model evaluated on an unbalanced dataset can be interpreted as the weighted sum of the errors evaluated on the different sky conditions. To detail the structural forecasting performance of the models, fifteen days corresponding to three distinct cloud conditions are isolated: clear-sky, broken-sky and overcast days (5 days each). We perform a quantitative and qualitative comparative analysis of the model predictions based on input data (SI: sky images, SO: satellite observations, IC: irradiance channels).

Table~\ref{tab:results_sky_conditions_all} presents a detailed analysis of weather and input specific performances. To begin with, the forecast accuracy is highly dependent on the type of weather conditions. The RMSE error ranges from around 20 to 60 W/$\text{m}^2$ in clear-sky days, 160 to 200 W/$\text{m}^2$ in broken-sky days and 50 to 135 W/$\text{m}^2$ in overcast days. In clear-sky conditions, the SPM outperforms DL models by a large margin in all configurations (sky images, satellite observations, irradiance channels) and for all lead times (10 to 60-min ahead). Similarly, the best performing DL model barely improves over the SPM in overcast conditions with a FS ranging from 4 to 7\% when fed with sky images and irradiance channels. In contrast, ECLIPSE's FS improves over the SPM by around 20 to 30\% in broken sky days.

Regarding the impact of the type of input on the performances, models trained on satellite observations alone seem to benefit the most from the additional irradiance channel. This is especially visible in overcast weather conditions with a significant FS gain of more than 70\%, hence closing the gap with models trained on sky images which are easier to correlate with the current irradiance level. Overall, the model trained with all three input types (sky images, satellite observations, irradiance channels) performs the best in clear-sky conditions up to a 50-min lead time, whereas the one trained with sky images and irradiance channels is the best in overcast conditions. For broken-sky days, the input setups including sky images lead to similar performances (26 to 29\% FS) with a slight difference between short-, medium- and long-term forecasts: the irradiance channel benefits shorter lead times the most, while training on sky images alone provides the most accurate 50 to 60-min ahead forecasts.

Prediction curves corresponding to three weather conditions (sunny, broken-sky and overcast) are used to illustrate some of the forecast properties for different types of input and forecast horizons. Figure~\ref{fig:clear_sky_day} shows the 30-min ahead predictions of the models over a clear-sky day (15/09/2019). The absence of the main source of variability in cloud-free days leads to little solar flux fluctuation. The corresponding deterministic irradiance changes are therefore easy to predict with the SPM. Overall, all models behave similarly showing smooth upward and downward predictions close to the ground truth at the beginning and at the end of the day. The highest source of errors appears to be when the clear-sky irradiance is the highest, which illustrates the difficulty for models to correlate an image with the corresponding irradiance level (9:00 to 14:00). During that time, the additional IC seems to benefit the model based on both sky and satellite images the most.

Figure~\ref{fig:broken_sky_day} presents the predictions of the models in broken-sky conditions, i.e. high irradiance variability (17/06/2019). All models follow a similar trend within a small 200 W/$\text{m}^2$ interval compared to about 800 W/$\text{m}^2$ for the measured irradiance series. This characteristic of the forecasts was observed in~\citet{palettaBenchmarkingDeepLearning2021c} for shorter-term predictions.

Figures~\ref{fig:overcast_day} and~\ref{fig:overcast_day_inertia} both illustrate predictions in fully cloudy conditions which correspond to low irradiance measurements well below the clear-sky irradiance. The most striking aspect of the forecasts depicted in Figure~\ref{fig:overcast_day} is a large error bias (from 300 to 450 W/$\text{m}^2$ on average) visible for the model predicting from satellite images only. Conditioning on past irradiance values significantly decreases this error to around 100 W/$\text{m}^2$. Surprisingly, adding an IC to both sky and satellite images raises this bias by a factor of two on average. In different overcast conditions, models suffer from a similar consistent bias (from noon in Figure~\ref{fig:overcast_day_inertia}). This could be caused by the difficulty in estimating the current level of irradiance or in limiting the risk of large errors caused by unpredicted upward irradiance shits. In addition, a strong inertia is visible in the predictions made by the model trained on sky images alone: both peaks measured around 8:20 and 10:20 (Ground truth), are predicted at the same time as the SPM, about one hour after the actual events. Long-term forecasts of models predicting from sky images only are indeed expected to face the persistence barrier - inability to foresee events before they happen, i.e. to decrease time lag below the forecast horizon (\citet{palettaECLIPSEEnvisioningCloud2021}) - as sky images offer less visibility into the future compared to satellite observations, which reveal distant clouds better.

\begin{table*}[ht!]
\caption{Relative advantages of sky and satellite images in vision-based intra-hour irradiance forecasting. The model is trained to predict future irradiance values from past sky images (SI), satellite observations (SO) and irradiance channels (IC). The $\alpha$ coefficient corresponding to the image prediction loss $L_{image}$ in Equation~\ref{equ:total_loss_with_distribution} is set to $0$.}
\vspace{-0.5\baselineskip}
\begin{center}
\begin{tabular}{lccccccccc}
\hline
\noalign{\vskip 1mm}
 & & & & \multicolumn{6}{c}{CRPS $\downarrow$ [W/$\text{m}^2$] (Forecast Skill $\uparrow$ [\%])}\\
 \multicolumn{3}{c}{Forecast Horizons} & $\mid$ & 10-min & 20-min & 30-min & 40-min & 50-min & 60-min \\
\hline\hline
\noalign{\vskip 2mm}
%Smart Pers. &&  (0\%) &  (0\%) &  (0\%) & &  &  &  \\
%\noalign{\vskip 1mm}
%ECLIPSE &&  &  &  & &  &  &  \\
%\noalign{\vskip 0.3mm}
 \multicolumn{3}{c}{Smart Pers.} && 73.2 (0\%) & 84.8 (0\%) & 91.2 (0\%) & 95.3 (0\%) & 96.3 (0\%) & 98.3 (0\%) \\
\noalign{\vskip 2mm}

SO & SI & IC & &  &  &  &  &  &  \\ %\checkmark

\checkmark &&&& 69.8 (4.6\%) & 70.3 (17.1\%) & 71.8 (21.3\%) & 72.6 (23.8\%) & 73.3 (23.9\%) & 73.8 (24.9\%) \\
\checkmark && \checkmark && 51.2 (30.0\%) & 56.4 (33.5\%) & 59.3 (35.0\%) & 61.0 (36.0\%) & 63.0 (34.5\%) & 65.8 (33.0\%) \\
\noalign{\vskip 2mm}
& \checkmark &&& 46.9 (36.0\%) & 52.0 (38.7\%) & 55.7 (38.9\%) & 57.8 (39.3\%) & 59.1 (38.7\%) & 60.2 (38.8\%) \\
& \checkmark & \checkmark && \textbf{44.4 (39.3\%)} & 51.0 (39.9\%) & 54.5 (40.2\%) & 56.4 (40.8\%) & 58.6 (39.1\%) & 61.9 (37.0\%) \\
\noalign{\vskip 2mm}
\checkmark & \checkmark &&& 48.4 (33.8\%) & 52.6 (38.0\%) & 55.1 (39.6\%) & 57.1 (40.1\%) & 58.2 (39.6\%) & \textbf{58.4 (40.6\%)} \\
\checkmark & \checkmark & \checkmark && 44.7 (38.9\%) & \textbf{50.6 (40.4\%)} & \textbf{53.9 (40.9\%)} & \textbf{55.8 (41.4\%)} & \textbf{57.1 (40.7\%)} & 58.8 (40.2\%) \\
\noalign{\vskip 1mm}
\hline
\end{tabular}
\end{center}
\label{tab:results_eclipse_crps}
\end{table*}

\subsection{Uncertainty quantification}
\label{uncertainty}

\begin{figure}%[H]%[ht!]%[h!] 
\centering
\includegraphics[width=0.45\textwidth]{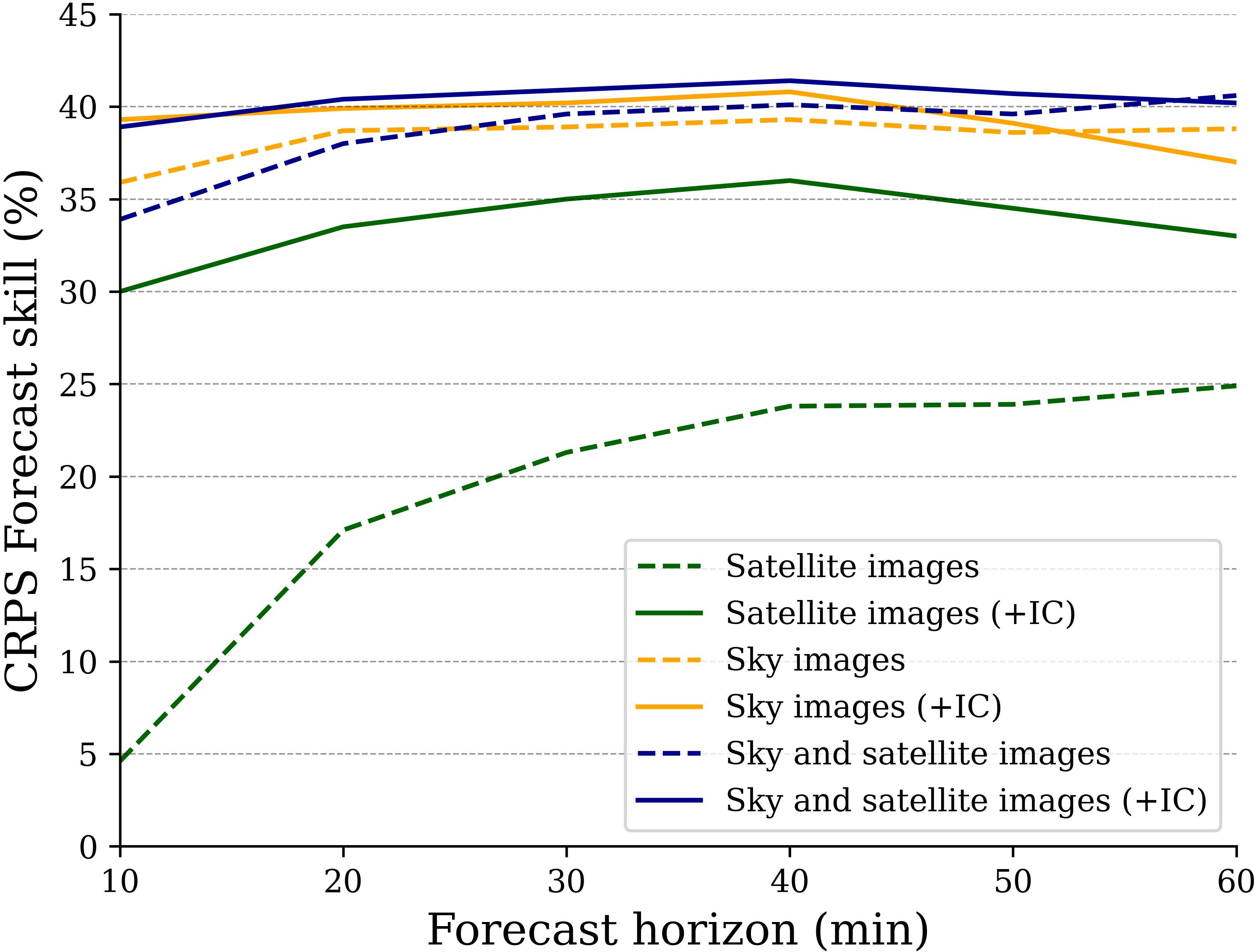}
%\vspace{-1.2\baselineskip}
\caption{Comparative performance of the different types of observation in probabilistic intra-hour irradiance forecasting using the FS score based on the CRPS metric. Without past irradiance conditioning (no IC), the gain of using satellite images in addition to sky images is visible from a 25-min lead time.}
\label{fig:horizons_crps}
\end{figure}

\begin{figure*}%[H]%[ht!]%[h!] 
\centering
\begin{minipage}[b]{0.48\textwidth}
    \includegraphics[width=1.0\textwidth]{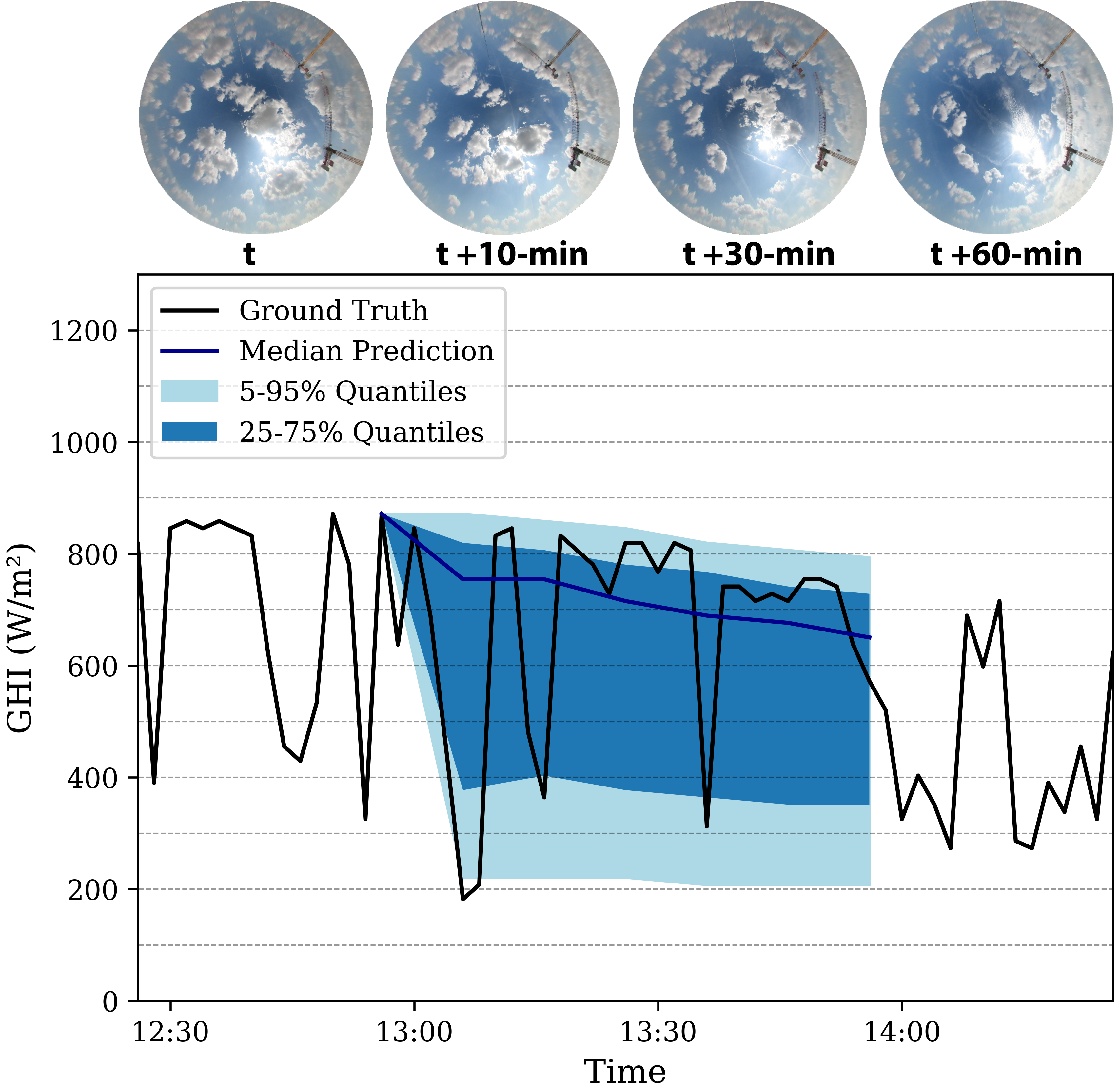}
  \end{minipage} 
  %\quad
  \begin{minipage}[b]{0.48\textwidth}
    \includegraphics[width=1.0\textwidth]{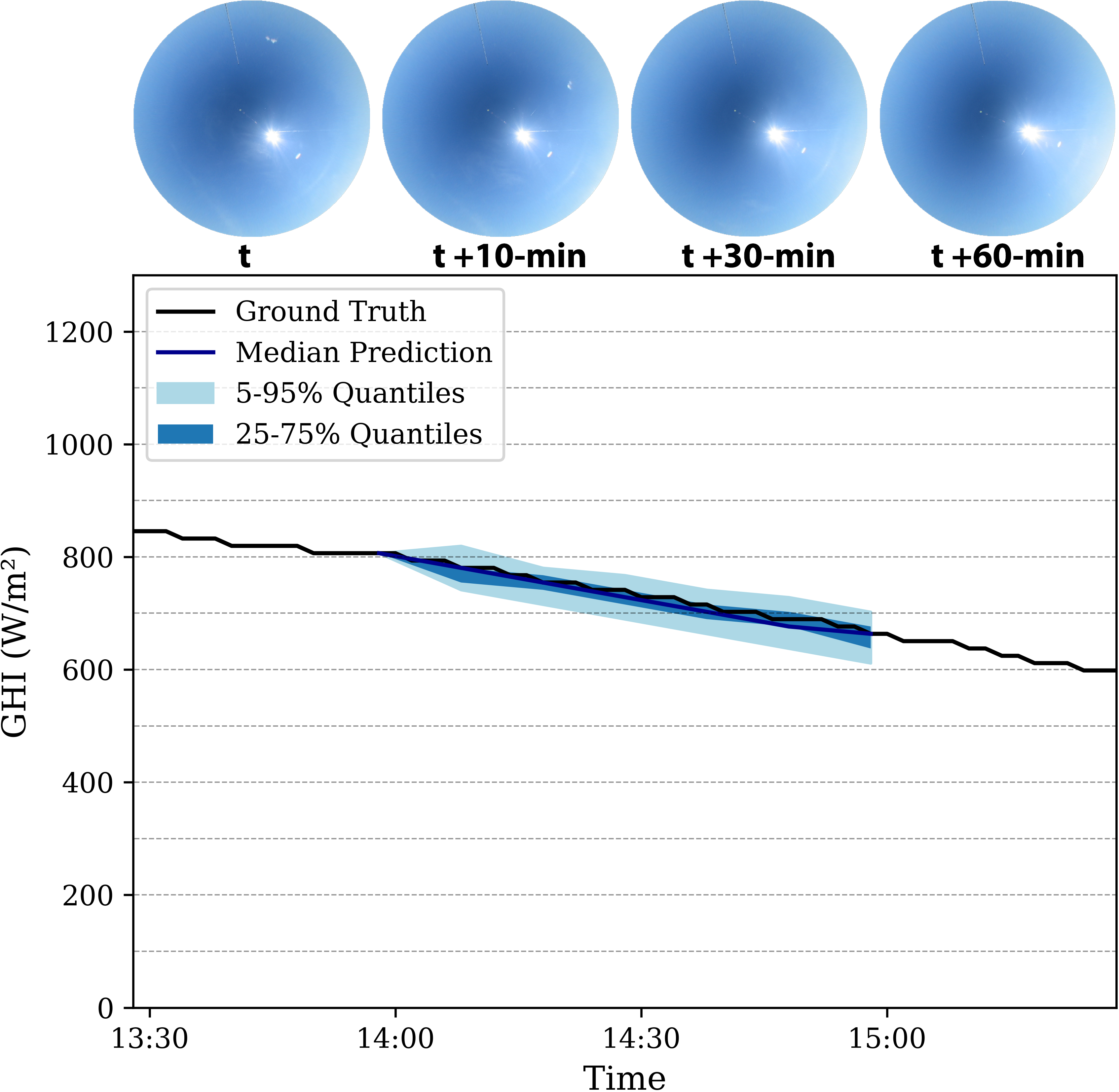}
\end{minipage}

\caption{Probabilistic prediction curves based on past sky and satellite images (no IC) in two different weather conditions: in the left panel, a broken sky context corresponding to highly variable solar irradiance levels (12/05/2019); in the right panel, a clear-sky day associated with low solar variability (13/05/2019). Despite similar median trends of the 10 to 60-min ahead forecasts, the associated uncertainties predicted by the model differ widely, the more variable the irradiance values, the wider the predicted distribution. In these examples, the probabilistic framework offers a richer modelling setup compared to deterministic predictions.}
\label{fig:uncertainties_extrems}
\end{figure*}

\begin{figure*}%[H]%[ht!]%[h!] 
\centering
\begin{minipage}[b]{0.48\textwidth}
    \includegraphics[width=1.0\textwidth]{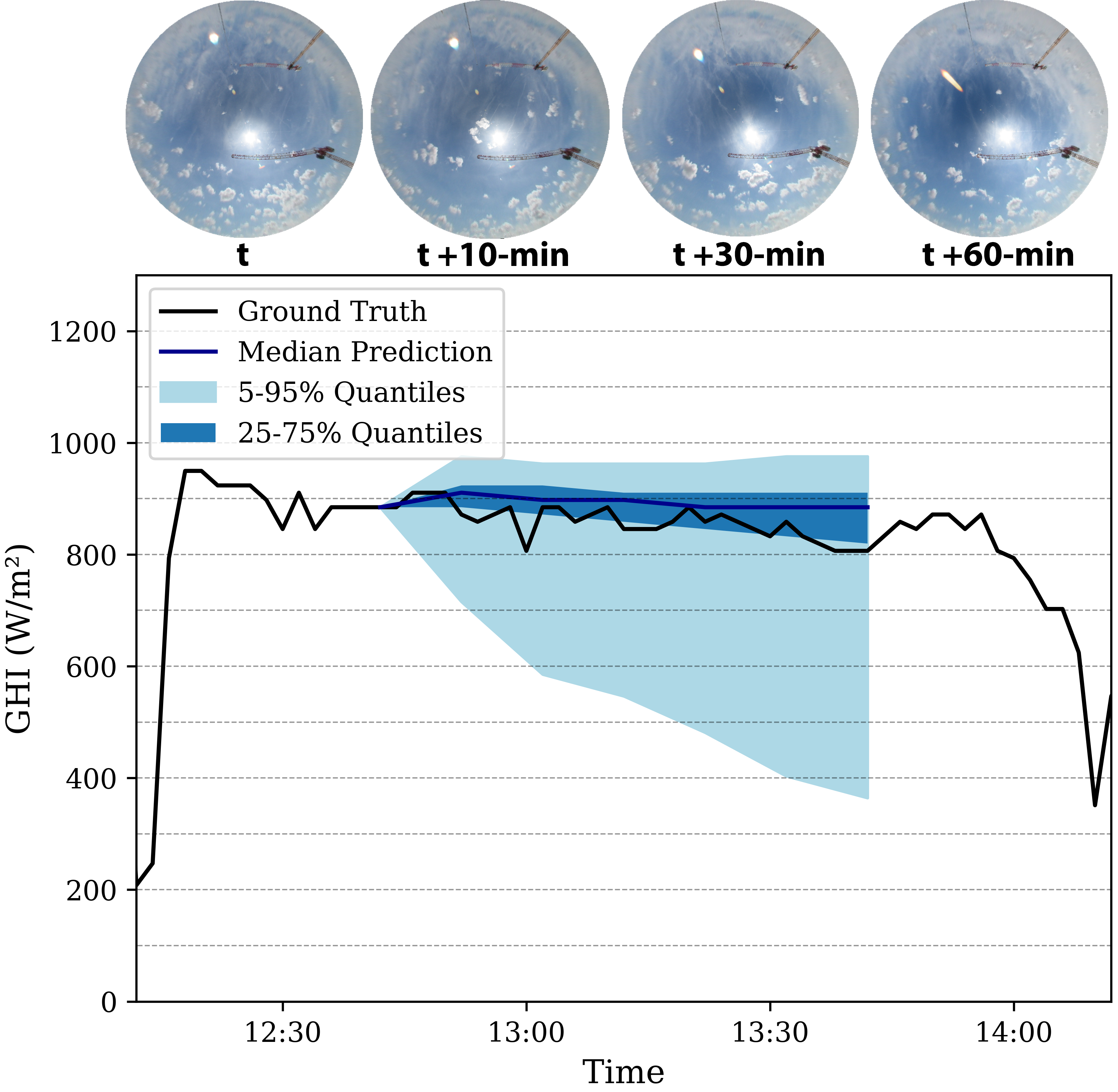}
  \end{minipage} 
  %\quad
  \begin{minipage}[b]{0.48\textwidth}
    \includegraphics[width=1.0\textwidth]{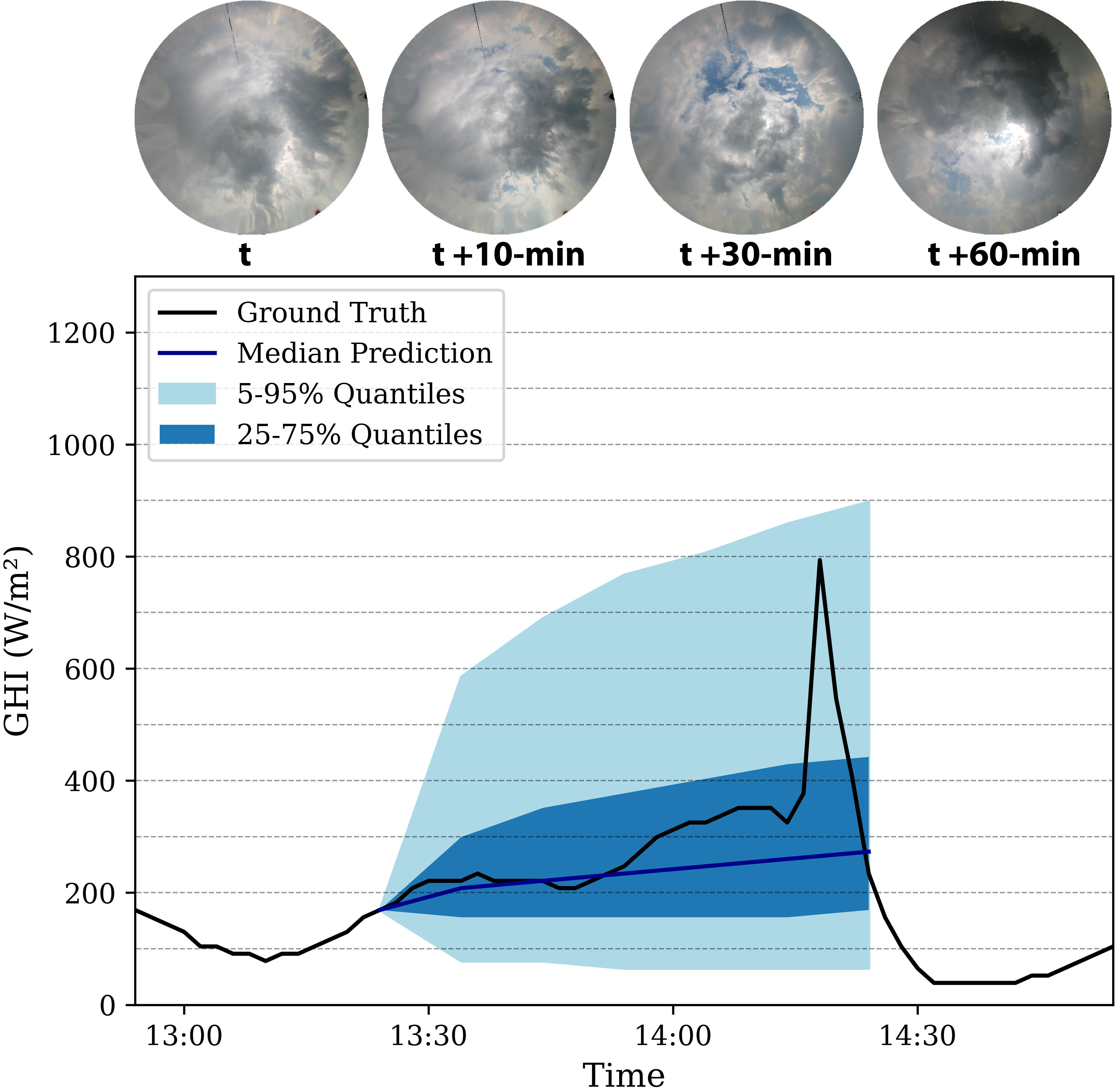}
\end{minipage}

\caption{Probabilistic prediction curves based on past irradiance values (IC), sky and satellite images in two different weather conditions: in the left panel, a clear-sky day corresponding to high solar irradiance levels (21/07/2019); in the right panel, an overcast day associated with low irradiance values (21/05/2019). In both cases, the predicted uncertainty increases with the horizon (light and dark blue areas). Noticeably, both forecasted distributions are asymmetric but spread out toward opposite irradiance values relative to the median trend: lower irradiance levels in mostly clear-sky conditions and higher irradiance levels in cloudy conditions. These correspond indeed to the most likely incoming irradiance fluctuations given past measurements and the current cloud cover conditions.}
\label{fig:uncertainties_examples}
\end{figure*}

Probabilistic solar irradiance forecasting with DL offers a richer modelling framework than single irradiance level prediction (\citet{carriereNewApproachSatelliteBased2021}). In particular, uncertainty quantification is valuable for many applications (hybrid power plant, network balance, electricity trading, etc.). In this section, a range of experiments are conducted to evaluate the impact of input data on predicted uncertainties. In addition, a few qualitative examples are used to illustrate some features of probabilistic irradiance forecasting with computer vision.

Table~\ref{tab:results_eclipse_crps} highlights experimental results obtained by training the model to predict future irradiance distributions from different data sources (sky and satellite images, irradiance channels). The CRPS metric used to evaluate probabilistic predictions shows that models using sky images or irradiance channels perform the best on average. Similarly to single point irradiance forecasting, adding the extra irradiance channel improves short-term forecasts the most. The configuration including all three data sources has the lowest CRPS on the 20 to 50-min ahead predictions on average (Figure~\ref{fig:horizons_crps}), whereas the model learning from sky images and past irradiance values provides the most accurate very short-term probabilistic forecast (10-min lead time).

To further analyse the differences in term of uncertainty predictions, four specific example forecasts are depicted in Figures~\ref{fig:uncertainties_extrems} and~\ref{fig:uncertainties_examples}. The first two examples correspond to two weather types, broken and clear-sky, associated with high and low irradiance variabilities respectively (Figure~\ref{fig:uncertainties_extrems}). Both predictions are based on the same input data (past sky and satellite images). Interestingly, in the context of high irradiance fluctuations (800 W/$\text{m}^2$ interval), the median forecast of the model remains relatively steady within a 150 W/$\text{m}^2$ interval, whereas the 5-95\% quantile area includes most irradiance highs and lows. By contrast, the second example corresponding to a steady solar flux decrease displays a very similar downward median irradiance forecast (150 W/$\text{m}^2$ interval), but its associated uncertainty is much lower (about 100 W/$\text{m}^2$ between 5 and 95\% quantiles compared to around 700 W/$\text{m}^2$ in the first example). These examples clearly show the advantage of probabilistic predictions compared to deterministic forecasts in integrating the model uncertainty into downstream applications. Figure~\ref{fig:uncertainties_examples} shows two intra-hour forecasts in clear-sky and fully cloudy conditions. As expected in these conditions, predicted uncertainties are asymmetric towards lower irradiance levels when the sky is cloud-free (left panel) and towards higher levels when the sky is overcast (right panel). In addition, the increasing surface of the blue areas shows a steady increase of the uncertainty with the horizon.

%Appendix~\ref{section:image_input} presents an additional comparative experiments showing some of the impacts of the input data on the probabilistic predictions. For instance, Figure~\ref{fig:uncertainties_comparisons} shows that adding the IC to the satellite images (first raw) reduces the model uncertainty. In addition, the IC can have a strong impact on the 

%In addition, the predictions show a clear increase of the uncertainty with the forecast horizon.

%In particular, we expect the model uncertainty to grow with the forecast horizon or the irradiance variability.

%\subsection{Sampling strategies}
%\label{sampling}

%\subsection{Averaging windows}
%\label{averaging}

%Compare 1-min vs 5-min average predictions

%\subsection{Best prediction per horizon or best average prediction}
%\label{predictions}

%Figure : FS vs horizons (1. best per horizon, 2. best average)
%Could go with temporal resolution of predictions

%\subsection{PCA}
%\label{pca}

\subsection{Predicted sky images}
\label{predicted_images}

\begin{figure*}%[H]%[ht!]%[h!] 
\centering    
\includegraphics[width=1\textwidth]{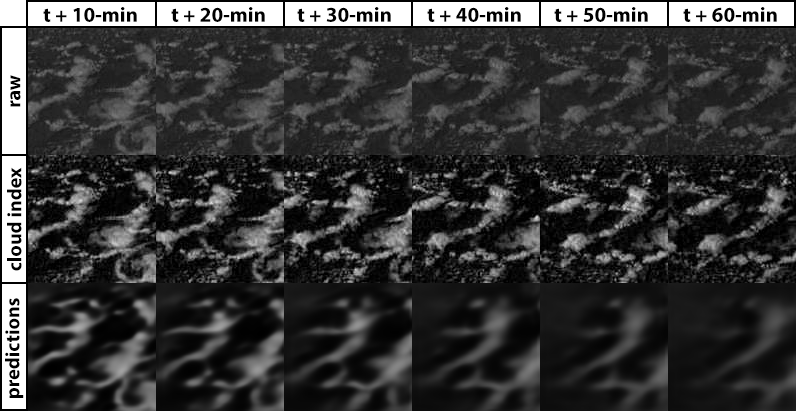}
\vspace{-0.5\baselineskip}
\caption{Prediction of future cloud index maps from past satellite observations (27/04/2019). The overall displacement of the cloud cover from left to right is well anticipated by the model. Notice how contrast decreases towards longer-term predictions and how clouds emerging from the left border of the image are not visible in output frames.}
\label{fig:video_prediction_1}
\end{figure*}

\begin{figure*}%[H]%[ht!]%[h!] 
\centering    
\includegraphics[width=1\textwidth]{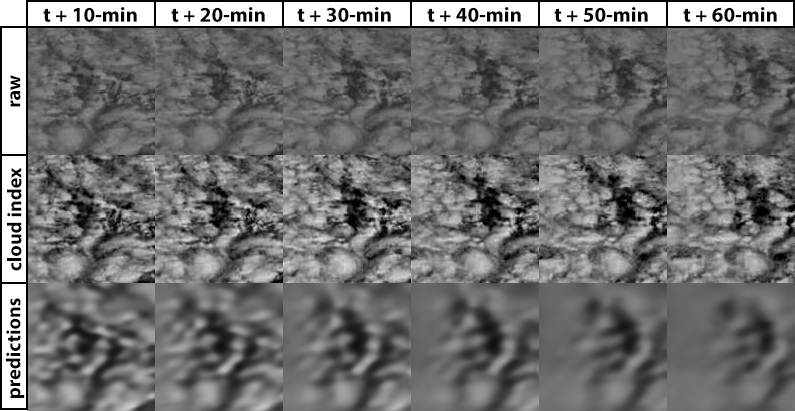}
\vspace{-0.5\baselineskip}
\caption{Prediction of future cloud index maps from past satellite observations (09/05/2019). Contrary to Figure~\ref{fig:video_prediction_1}, the model rightly predicts that clouds are going to appear from the left border.}
\label{fig:video_prediction_2}
\end{figure*}

The prediction of future cloud index maps in addition to irradiance levels not only improves the forecasting performance of a DL model but it also illustrates its ability to accurately model cloud movements. Two examples are depicted in Figures~\ref{fig:video_prediction_1} and~\ref{fig:video_prediction_2}. The coefficient $\alpha$ is set to 1000 instead of 5 such that the video prediction task prevails over the irradiance forecasting task during training, hence showing the full capacity of the model in video prediction. In both cases the overall pattern displacement is well predicted despite fuzzier long-term predictions. Note how the model has to infer the part of the cloud coverage, which was not yet visible in the past observations.

\section{Discussion of limitations}
\label{discussion}

One of the limitations of the study is the integration in the modelling of different temporal resolutions, from 2-min for sky images to 5-min for satellite images. To increase the size of the training set which is key in vision-based irradiance forecasting (\citet{palettaBenchmarkingDeepLearning2021c}), the sequence of satellite images was not consistently temporally aligned with the sequence of sky images (see Section~\ref{irregular_sampling}). Although there is an advantage in terms of number of samples, the temporal shift up to 4-min of the sequence of satellite images relative to the last sky image leads to an additional uncertainty for the model, especially for shorter-term forecasting. Neural ordinary differential equations might offer a more appropriate framework for solar applications using irregularly-sampled data (\citet{chenNeuralOrdinaryDifferential2019}).

Another limitation of the proposed approach is the reliance on local observations (sky images and irradiance measurements), which might require site specific model training or fine tuning of a trained model with local observations. In the latter setting, the use of the polar coordinates (SPIN) or image rotations during training will likely facilitate knowledge sharing between different datasets by leveraging the rotational invariance of this computer vision problem as discussed in \citet{palettaSPIN2021}.

Finally, experiments with the IC have shown that models trained on satellite images and their corresponding effective cloud albedo seem to lack information on the current state of the solar radiation. The difficulty of correlating an image with a level of solar irradiance would benefit from additional preprocessing of the satellite image. Using known physics of the interaction between solar flux and clouds to base predictions on irradiance maps instead of cloud index maps might significantly improve predictions which are not relying on past irradiance measurements.

\section{Conclusion}
\label{conclusion}

Intra-hour irradiance forecasting is key to facilitating the integration of solar into the energy mix. Vision-based approaches based on deep learning models aim at providing statistically realistic forecasts based on training and input data but also anticipating future irradiance fluctuations based on the current cloud cover dynamics. This study bridges the gap between solar energy forecasting from sky images and from satellite observations, by proposing a deep learning architecture with parallel encoders to predict future irradiance levels from both types of images.

This hybrid model is shown to outperform the standard models trained on sky or satellite images only from the 25-min forecast horizon. Regarding sky conditions, the hybrid model provides the most accurate predictions in clear-sky days, whereas the model learning from sky images is the best in overcast  and broken-sky weathers. In contrast, the model predicting irradiance from satellite images is strongly penalised by the problems in correlating satellite observations with their corresponding irradiance levels. Thus, it significantly benefits from conditioning predictions on past irradiance measurements by the addition of an extra irradiance channel.

Finally, probabilistic predictions demonstrate a richer forecasting framework by facilitating uncertainty quantification in cloudy conditions and for long-term predictions. We hope that this research will foster interest in combining sky images with satellite observations for intra-hour solar irradiance forecasting.

\vspace{1\baselineskip}
{\bf Acknowledgements} The authors acknowledge SIRTA and EUMETSAT for providing the data used in this study. We are grateful to Aleksandra Marconi, Anthony Hu and Marcos Gomes-Borges for their technical assistance and valuable comments on the manuscript. This research was supported by ENGIE Lab CRIGEN, EPSRC and the University of Cambridge.

%\bibliographystyle{elsarticle-harv} 
%\bibliographystyle{elsarticle-num} 
%\bibliographystyle{elsarticle-num-names} 
%\bibliography{library.bib}

\bibliographystyle{elsarticle-harv} 
\bibliography{library}

%% The Appendices part is started with the command \appendix;
%% appendix sections are then done as normal sections
%% \appendix

%% \section{}
%% \label{}

\newpage

\appendix

\onecolumn

\renewcommand\thefigure{\thesection.\arabic{figure}}
\renewcommand\thetable{\thesection.\arabic{table}}

\section{Dataset balance}
\label{section:dataset_balance}
\setcounter{figure}{0}
\setcounter{table}{0}

\begin{figure}[h!]%[H]%[ht!]%[h!] 
\centering
\begin{minipage}[b]{0.32\textwidth}
    \includegraphics[width=1\textwidth]{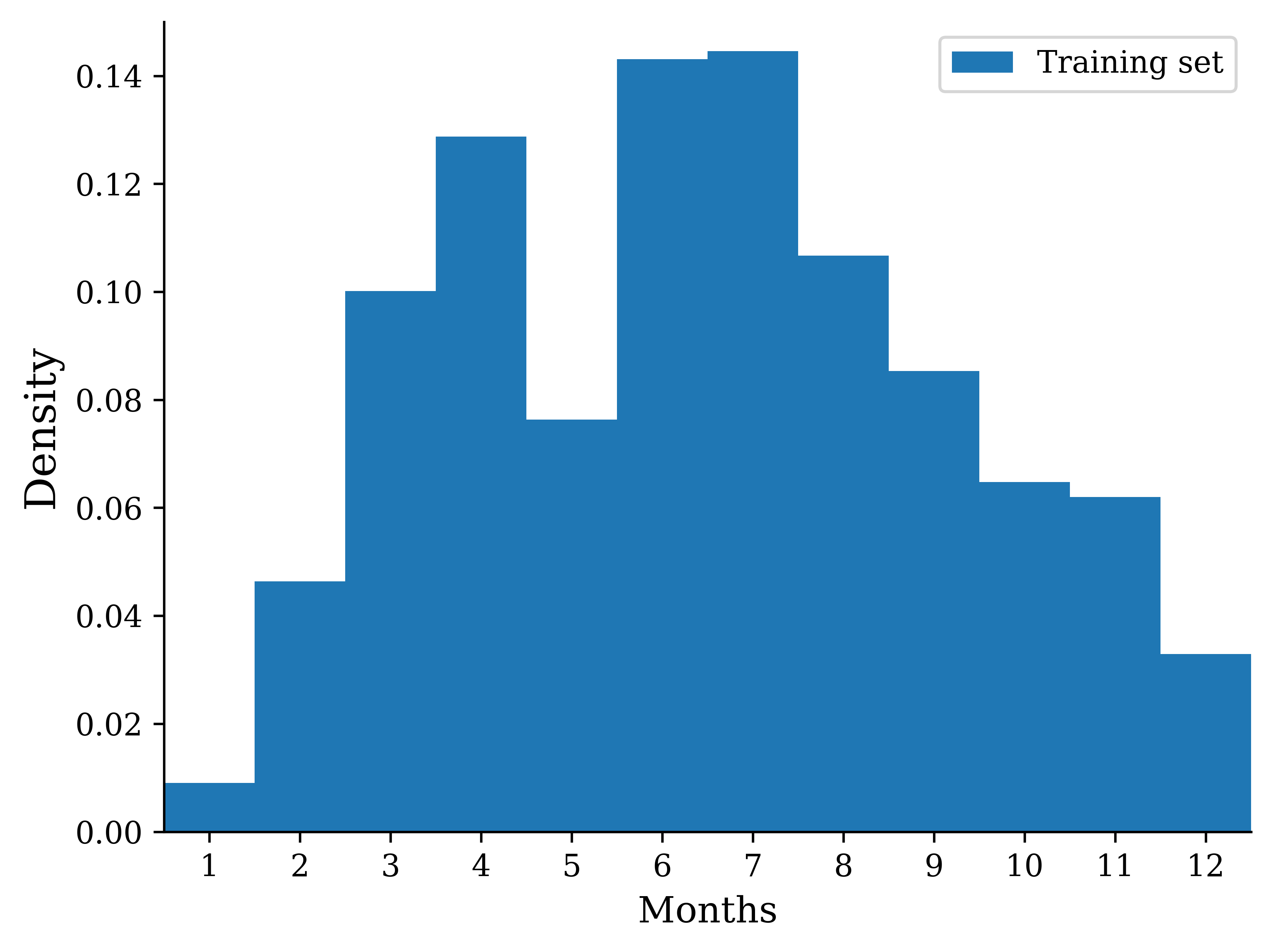}
    \label{fig:hist_months_train}
  \end{minipage} 
  \begin{minipage}[b]{0.32\textwidth}
    \includegraphics[width=1\textwidth]{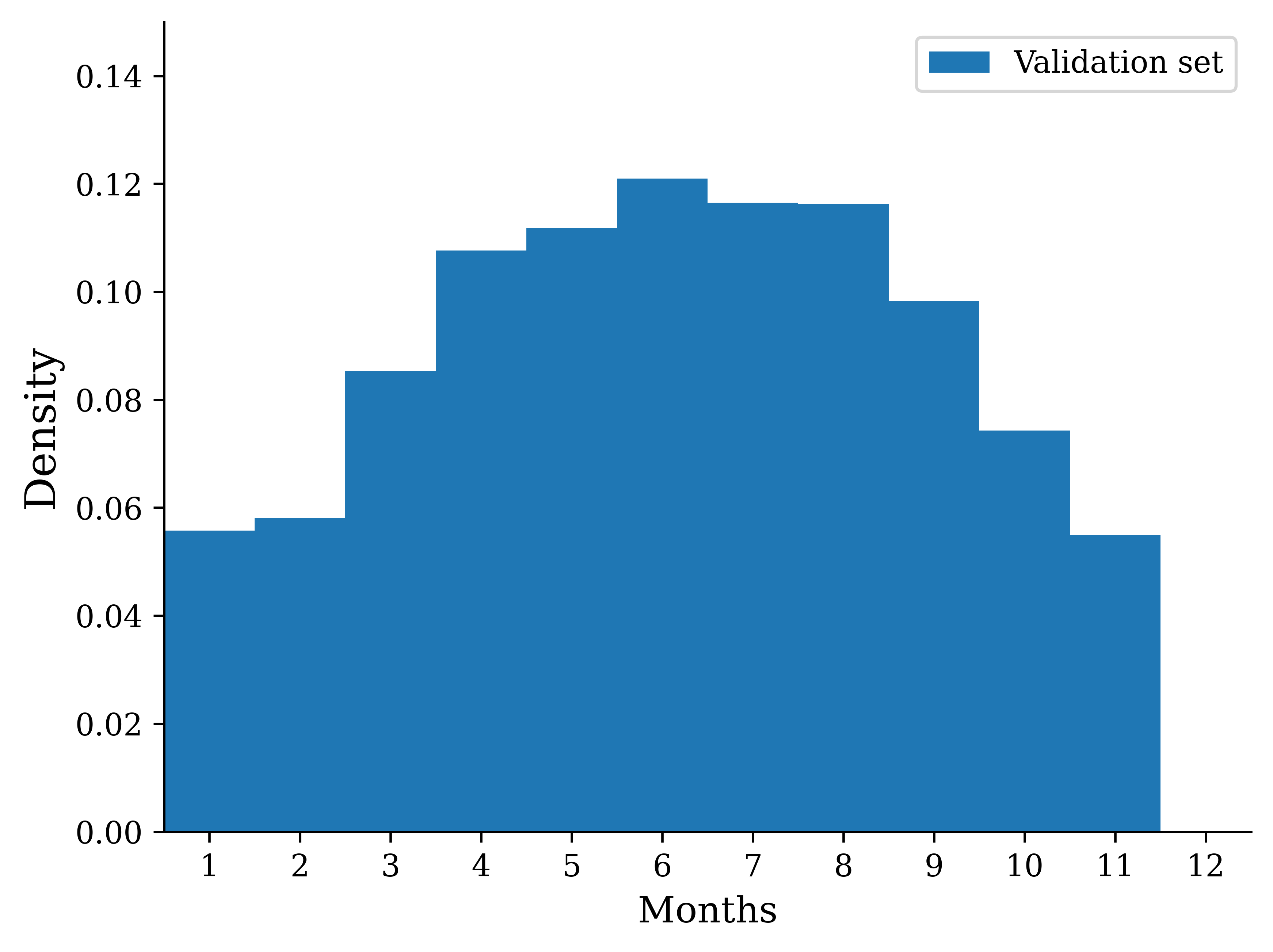}
    \label{fig:hist_months_val}
  \end{minipage}
  \begin{minipage}[b]{0.32\textwidth}
    \includegraphics[width=1\textwidth]{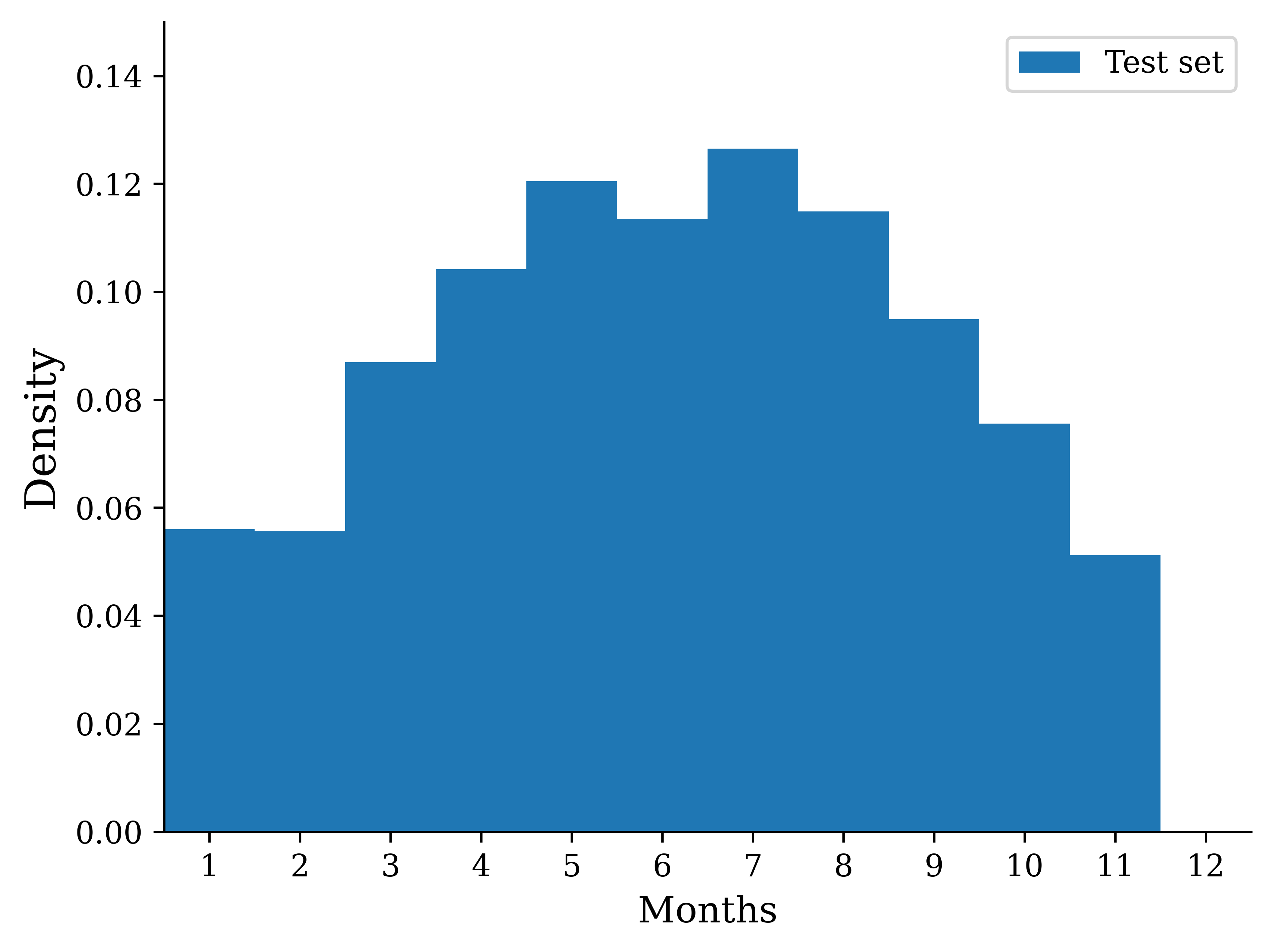}
    \label{fig:hist_months_test}
  \end{minipage}
  \vspace{-1\baselineskip}
\caption{Distribution of samples by months in the training, validation and test sets.}
\label{fig:dataset_distribution_month}
\end{figure}

  \vspace{-1\baselineskip}

\begin{figure}[h!]%[H]%[ht!]%[h!] 
\centering
\begin{minipage}[b]{0.32\textwidth}
    \includegraphics[width=1\textwidth]{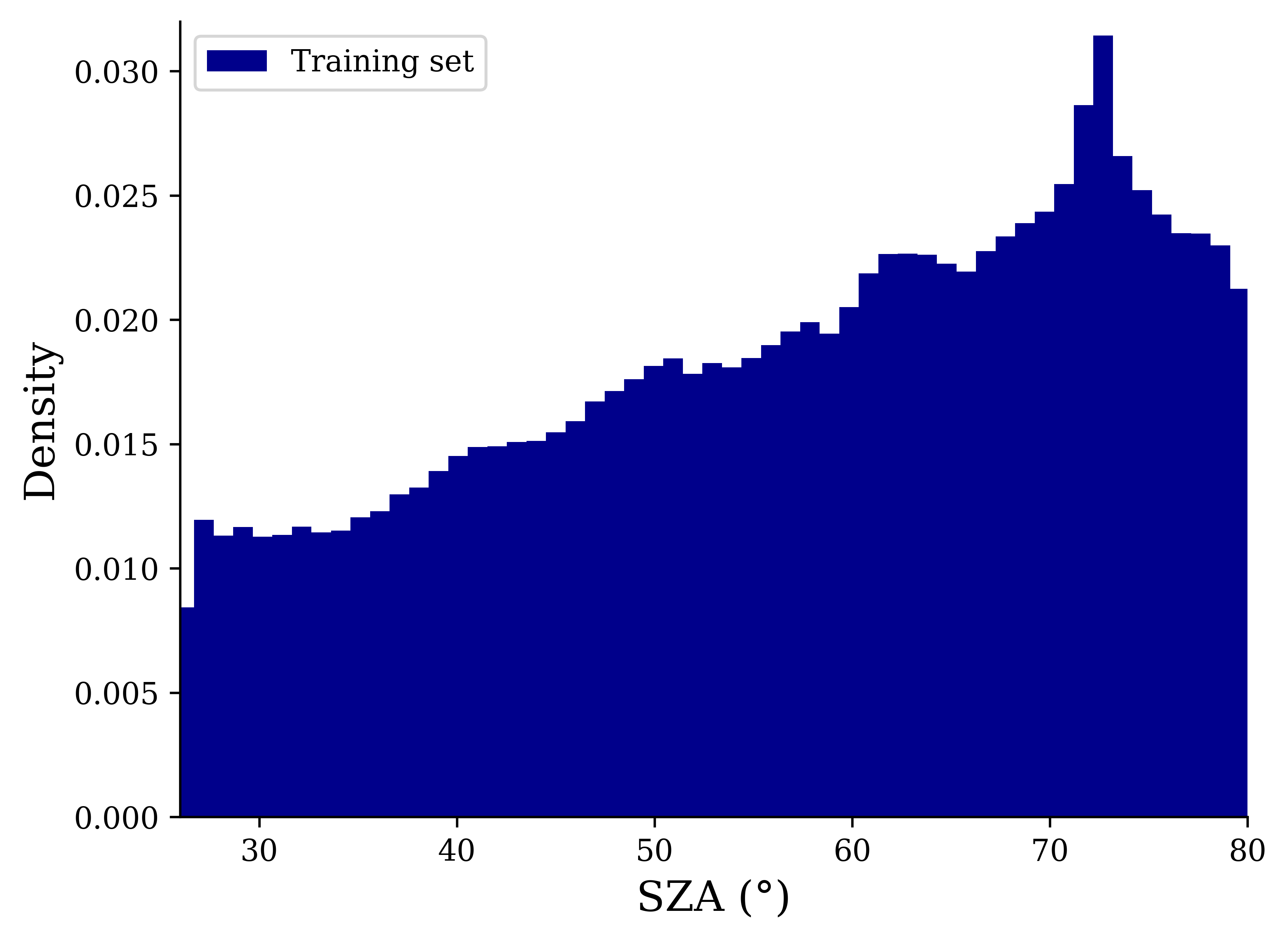}
    \label{fig:hist_sza_train}
  \end{minipage} 
  \begin{minipage}[b]{0.32\textwidth}
    \includegraphics[width=1\textwidth]{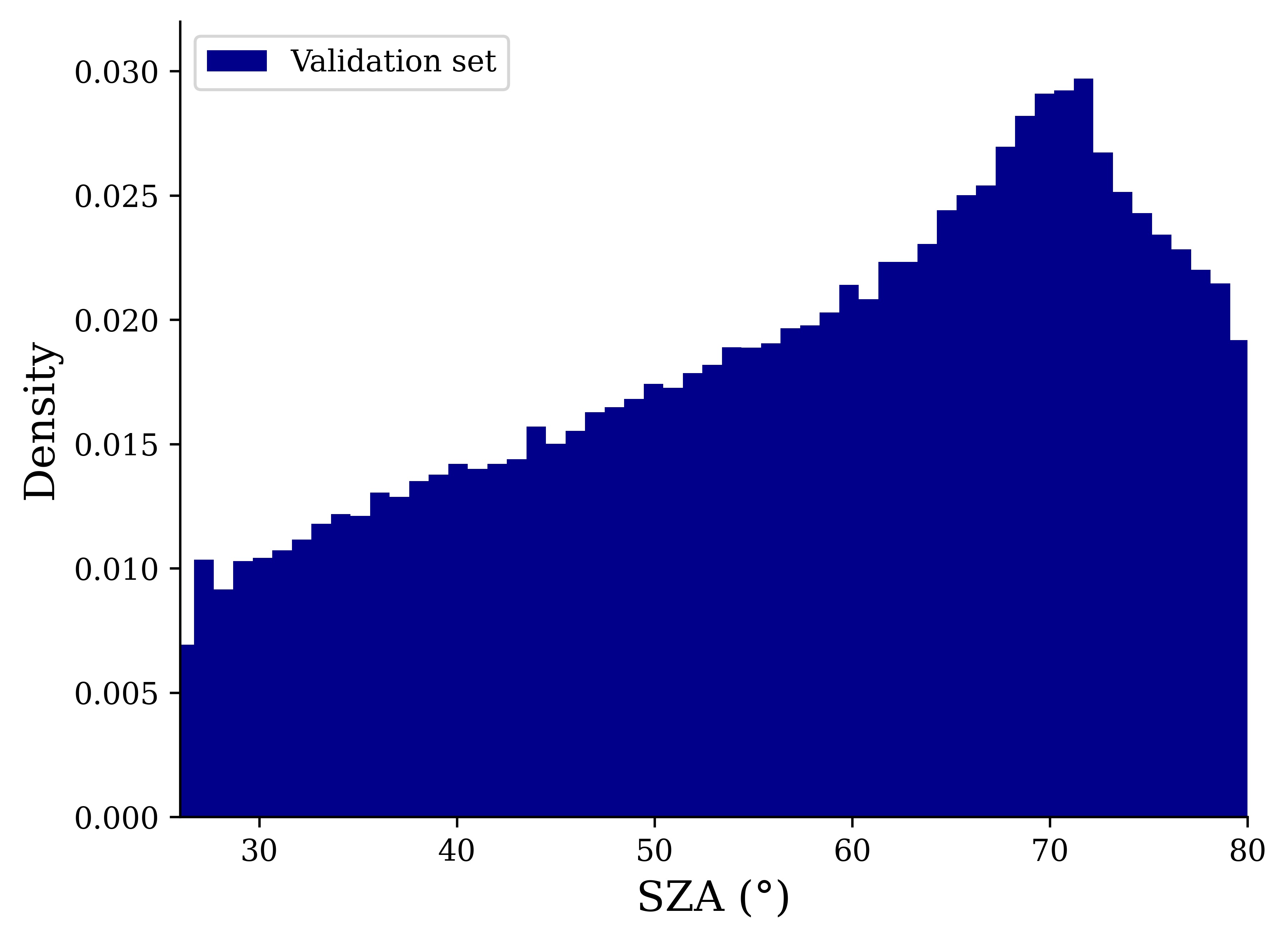}
    \label{fig:hist_sza_val}
  \end{minipage}
  \begin{minipage}[b]{0.32\textwidth}
    \includegraphics[width=1\textwidth]{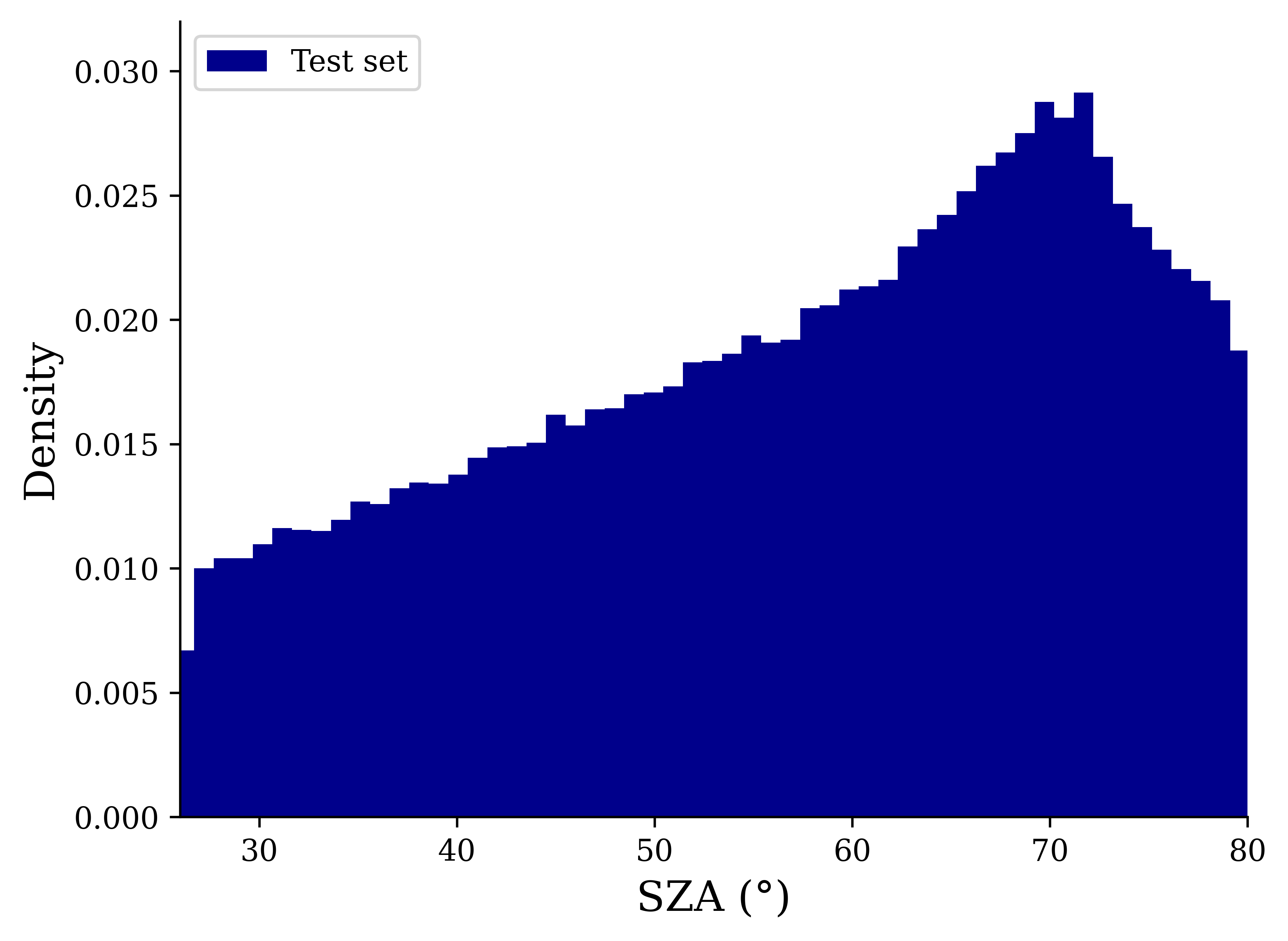}
    \label{fig:hist_sza_test}
  \end{minipage}
\vspace{-1\baselineskip}
\caption{Distribution of samples by Solar Zenith Angle in the training, validation and test sets.}
\label{fig:dataset_distribution_sza}
\end{figure}

\vspace{1\baselineskip}

\begin{figure}[h!]%[H]%[ht!]%[h!] 
\centering
\begin{minipage}[b]{0.32\textwidth}
    \includegraphics[width=1\textwidth]{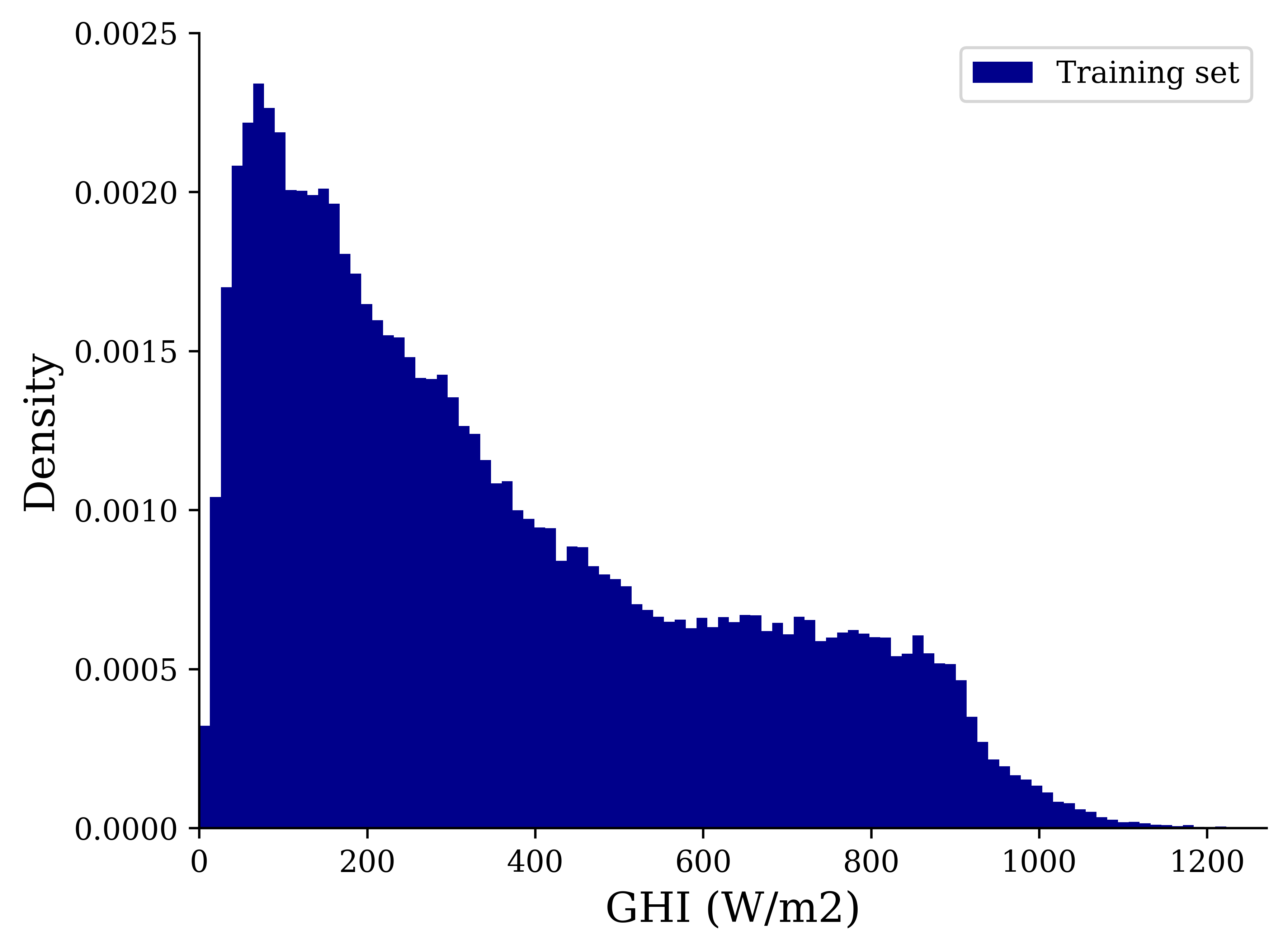}
    \label{fig:hist_bins_train}
  \end{minipage} 
  \begin{minipage}[b]{0.32\textwidth}
    \includegraphics[width=1\textwidth]{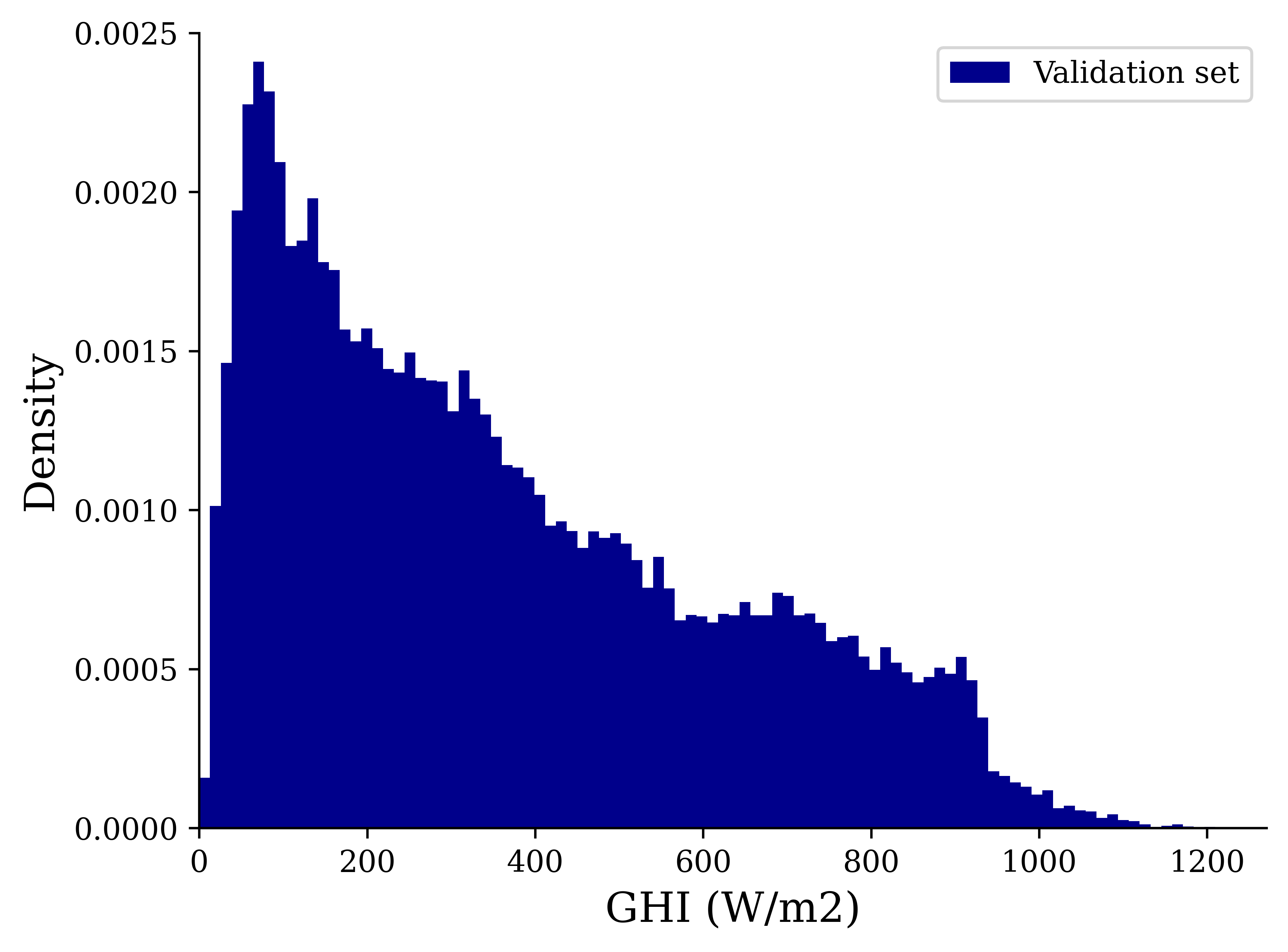}
    \label{fig:hist_bins_val}
  \end{minipage}
  \begin{minipage}[b]{0.32\textwidth}
    \includegraphics[width=1\textwidth]{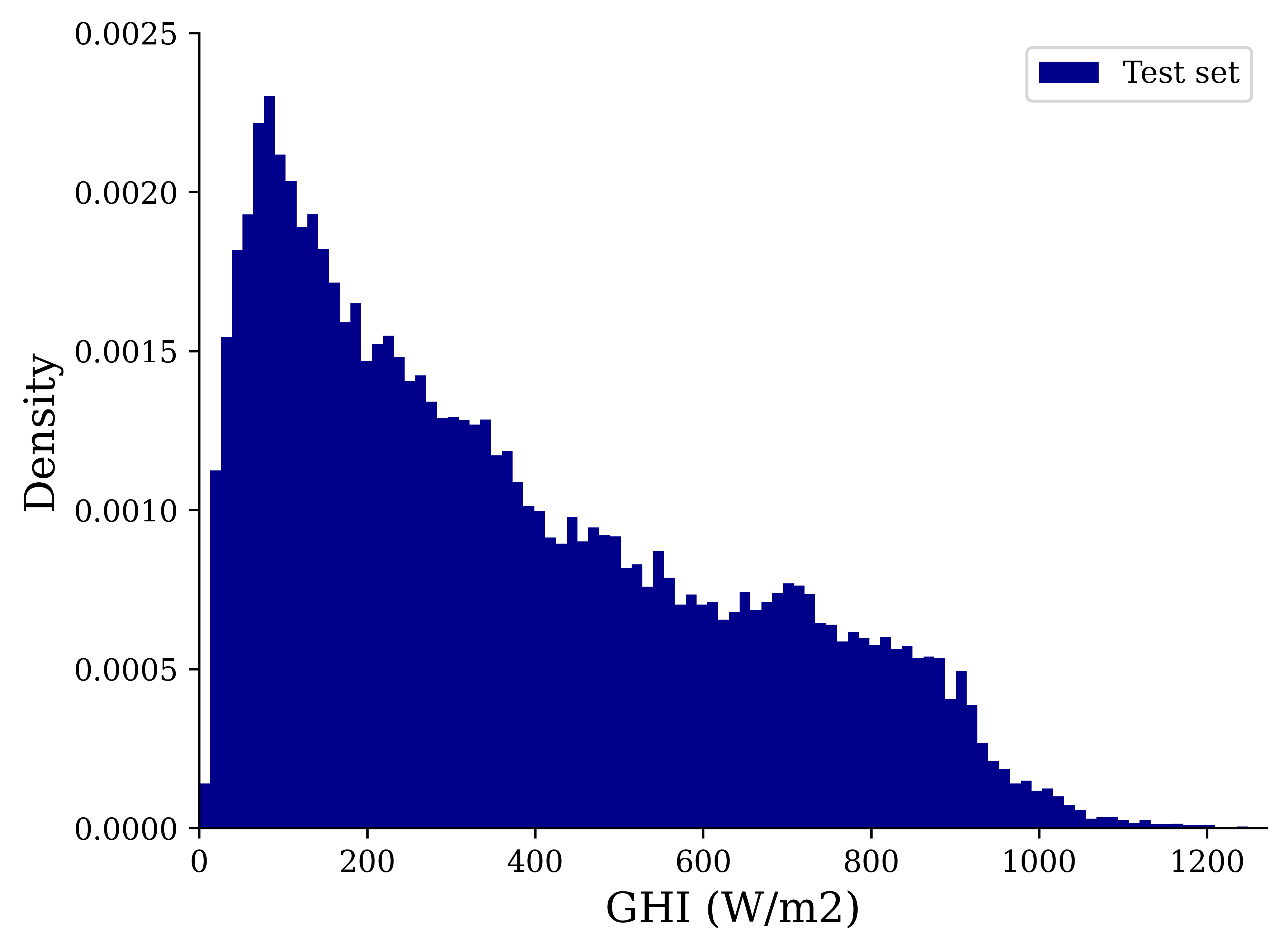}
    \label{fig:hist_bins_test}
  \end{minipage}
  \vspace{-1\baselineskip}
\caption{Distribution of samples by Global Horizontal Irradiance level in the training, validation and test sets.}
\label{fig:dataset_distribution_bins}
\end{figure}

%\break
%\clearpage
\newpage
\end{document}